\documentclass[11pt]{article}

\usepackage[final]{acl}

\usepackage{times}
\usepackage{latexsym}

\usepackage[T1]{fontenc}

\usepackage[utf8]{inputenc}

\usepackage{microtype}

\usepackage{inconsolata}

\usepackage{graphicx}

\usepackage{enumitem}
\usepackage{amsmath}
\usepackage{tabularx}
\usepackage{float}
\usepackage{multirow}
\usepackage{longtable}
\usepackage{hyperref}
\usepackage{stfloats}
\usepackage{booktabs}
\usepackage{adjustbox}

\title{From Global to Local: Learning Context-Aware Graph Representations for Document Classification and Summarization}

\author{
\textbf{Ruangrin Ldallitsakool}\textsuperscript{1}
\and
\textbf{Margarita Bugue{\~n}o}\textsuperscript{1,2}
\and
\textbf{Gerard de Melo}\textsuperscript{1,2}
\\
\textsuperscript{1}University of Potsdam
\quad
\textsuperscript{2}Hasso Plattner Institute (HPI)
\\
\texttt{ldallitsakool@uni-potsdam.de} \\
\texttt{\{margarita.bugueno, gerard.demelo\}@hpi.de}
}

\begin{document}
\maketitle
\begin{abstract}
Recent NLP systems commonly represent documents as linear token sequences. Although this captures sequential order, it can hinder modeling long-range dependencies and global document structure, especially for long texts. 
This paper proposes a data-driven method to automatically construct graph-based document representations. 
Building upon the recent work of \citet{bugueno2025}, we leverage the dynamic sliding-window attention module to effectively capture local and mid-range semantic dependencies between sentences, as well as structural relations within documents.
Graph Attention Networks (GATs) trained on our learned graphs achieve competitive results on document classification while requiring lower computational resources than previous approaches.
We further present an exploratory evaluation of the proposed graph construction method for extractive document summarization, highlighting both its potential and current limitations.
The implementation of this project can be found on GitHub\footnote{Project repository: \url{https://github.com/idalr/SlidingWindowAttnGraphs}}.
\end{abstract}

\section{Introduction}

In recent natural language processing (NLP) systems, documents are typically processed by language models (LMs) as linear sequences of tokens--such as characters, subwords, words, or sentences. While this sequential formulation allows models to capture token order, it often constrains their ability to represent long-range dependencies and higher-level document structures, particularly as input length increases \citep{hudson2022muld}. 
Moreover, LMs frequently encounter redundancy, where similar or identical content recurs within or across documents \citep{ma2022multi}, leading to inefficiencies in learning and representation as the model repeatedly encodes overlapping information.

One promising direction for effective and efficient document modeling lies in graph-based representations, where tokens are represented as nodes and edges explicitly capture structural relationships between node pairs.
Tokens referring to the same entity or exhibiting strong semantic similarity can be merged into a single shared node, enhancing representational efficiency, while edges may encode positional, semantic, or syntactic relations to capture both the document's structural information and contextual dependencies.

Most prior work relies on heuristically constructed graph-based document representations, where node and edge features are manually designed based on task objectives or domain expertise, such as linguistic knowledge.
However, designing effective graph representations remains challenging, as such approaches often struggle with domain transferability and depend heavily on external preprocessing steps \citep{bugueno2023connecting,wang2024graph}.

A complementary direction is to learn graph structure directly from data. Building on the graph structure learning framework proposed by \citet{bugueno2025}, this paper further advances automatically induced graph representations from attention models, with particular emphasis on improving \textbf{local context awareness} and \textbf{computational efficiency}. 

The main contributions of this paper are threefold:
(i) we extend the graph structure learning framework of \citet{bugueno2025} by replacing full attention with sliding-window attention, substantially reducing computational requirements;
(ii) we systematically evaluate multiple model configurations on three document classification benchmarks, achieving improved performance over prior results;
and (iii) we study the applicability of our approach to extractive document summarization, a task that has been largely underexplored in previous work.

\section{Related Work}

\subsection{Attention Mechanisms}

To overcome the challenges in capturing semantic and structural dependencies in long document processing, state-of-the-art models such as Transformers \cite{waswani2017attention} have been extended.
Notably, \citet{beltagy2020longformer} and \citet{zaheer2020big} introduced sparse attention mechanisms that extend the effective input length of Transformers while maintaining linear computational complexity.
Their models, \textit{Longformer} and \textit{BigBird}, demonstrated superior results across a range of NLP tasks.

Several sparse attention patterns have been proposed in their work.
\textbf{Sliding window attention} employs a fixed-size window around each token. It was proposed to reduce computation, as well as an effective tool for context localization \cite{kovaleva2019revealing}.
To further support mid-range context dependency, \citet{beltagy2020longformer} adapted \textbf{dilated sliding window attention} from dilated CNNs \cite{van2016wavenet}, to increase the window size without increasing the computational need.
\citet{zaheer2020big} introduced \textbf{random attention}, which allows their model to expand and approximate the different contexts.
Both works incorporate \textbf{global attention}, which aggregates the representation of a document on specially added or pre-selected tokens, to capture structural information and long-distance dependencies.

\subsection{Traditional Graph-Based Representations}

Graph models offer several advantages over sequential language models \citep{xu2021contrastive, bugueno2023connecting, wang2023docgraphlm}.
Earliest approaches, such as TextRank (\citealp{mihalcea2004textrank}; \citealp{hassan2007random}) and LexRank \cite{erkan2004lexrank}, represented words as nodes and their co-occurrences as edges. However, they lacked deep semantic structure despite a densely connected network \cite{bugueno2023connecting}.

Other approaches have designed document graphs that integrate both linguistic and semantic knowledge while exploring multiple levels of granularity. Nodes may represent linguistic units--such as morphemes, words, phrases, sentences, or entire documents\citep{qian2018graphie, yao2019graph}--or semantic units, including entities, keywords, or events \citep{zeng2020double, hua2023improving}.
Correspondingly, edges encode structural or semantic relationships among these units.
Some work (e.g., \citet{wang2023docgraphlm}) further introduce multiple types of nodes and edges to facilitate coexisting diverse forms of information propagation.

Nevertheless, heavily tailored graph representations are not necessarily beneficial.
On text classification tasks, simple word-level co-occurrence graphs proved to be sufficient \cite{castillo2017text}.
Notably, such simpler structures tend to yield better performance on longer documents while being more computationally efficient \cite{bugueno2023connecting}.
This may be because complex or noisy graphs hinder effective learning, even for specialized graph-based learning models.
Assuming dense and long-range relations between nodes could benefit a global view, but it is not necessarily optimal for learning.
Excessive connectivity could lead to \textit{oversmoothing}, where node representations become indistinguishable after iterative neighborhood aggregation \citep{zhang2020every,plenz2024graph}.

Moreover, heuristic-based graph constructions often rely on feature selection, which might not generalize well in cross-domain or cross-task settings.
Also, extracting features from documents may rely on external processing steps, e.g., part-of-speech tagging, named entity recognition, and coreference resolution, which can introduce cascading errors throughout the learning process.

\subsection{Combining Graph and LMs}

Recent work has explored hybrid approaches that combine textual and structural representations. These methods integrate the rich contextual understanding of LMs with the relational reasoning capabilities of graphs to improve performance across various NLP tasks.

ConTextING \cite{huang2022contexting} unified pretrained BERT embeddings with inductive graph neural networks (GNNs) to jointly learn document-level context and fine-grained word interactions via a sub-word graph, highlighting contextual word semantics and graph-based reasoning.

Similarly, \citet{onan2023hierarchical} presented a hierarchical graph-based framework by combining linguistic features, domain-specific graph construction, and contextual embeddings. The model integrates multi-level graph learning and attention mechanisms to capture complex structural relationships, and a dynamic fusion layer that combines these with BERT representations.

\citet{plenz2024graph} introduced the Graph Language Model (GLM), which jointly encodes text and graph triplets, enabling unified reasoning over linguistic and structural information while retaining the expressive contextual representations of language models.

\subsection{Learned Graph Representations}
Despite recent advances, most graph-based representations still depend on heuristic or manually engineered construction methods. 
Consequently, they may not generalize well across diverse domains nor be able to handle modern document processing challenges, such as capturing long-distance and non-linear dependencies, or managing imbalanced or lengthy document inputs.

To overcome these limitations, \citet{xu2021contrastive} proposed a hybrid approach that trains a Graph Attention Network (GAT) \cite{velivckovic2017graph} on passage embeddings in each document and employs contrastive learning to refine document representations in an unsupervised manner.

\citet{buterez2025end} proposed EAGLE, a pure attention-based architecture, consisted of an encoder and an attention pooling mechanism which eliminating the need for explicit message passing. Their results provided strong evidence that attention could capture rich relational structure without handcrafted aggregation operators.

GraphLSS \cite{bugueno2025graphlss} proposed automatically constructible heterogeneous graph representations by integrating lexical, structural and semantic information, resulting in a more interpretable and efficient alternative.

Drawing on these ideas, \citet{bugueno2025} proposed a fully data-driven approach that constructs graphs directly from raw text, eliminating the need for handcrafted construction rules and domain dependency.
Their method leveraged pre-trained language models and full self-attention mechanisms to model contextual relationships between sentences, enabling the automatic induction of meaningful graph structures.
To enhance sparsity and efficiency, they applied statistical filtering to retain only salient relationships, yielding sparse yet informative graphs.
Results showed that models trained on these learned attention-induced graphs consistently outperformed heuristic-based counterparts, highlighting both the robustness and adaptability of this approach while reducing manual and domain-specific dependencies.

However, while \citet{bugueno2025} effectively captured global contextual dependencies, their reliance on full attention mechanisms introduces quadratic computational complexity.
Moreover, global attention is not always necessary, especially in long documents, as distant sentences may share little semantic relevance.
Prior research, such as \citet{beltagy2020longformer} and \citet{zaheer2020big} showed that local and mid-range context modeling can capture most of the essential information with substantially lower resource demands.

Based on these insights, this work investigates how graph-based representations can more effectively and efficiently exploit localized contextual dependencies while maintaining strong semantic coherence.

\section{Method}
Overall, our methodology closely follows \citet{bugueno2025}'s procedure.
However, we replace the full-attention modules with sliding-window attention (SWA).
As implemented in models such as Longformer \cite{beltagy2020longformer}, sliding-window attention restricts each sentence’s attention to a local neighborhood.
It enhances sensitivity to local and mid-range contextual relationships while substantially improving computational efficiency and robustness.

The window size, therefore, plays a critical role in balancing contextual coverage and noise: smaller windows emphasize local coherence but limit global context, whereas larger windows capture broader dependencies at the risk of introducing irrelevant information and increased computational overhead.

While \citet{bugueno2025graphlss} report optimal performance using a window size of 40\%--defined by the number of sentences in each document, and based on sentence similarity--we hypothesize that learned attention weights provide a more effective way to identify task-relevant context. This allows the use of substantially smaller windows without degrading performance.
Moreover, the dynamic adjustment to the input size allows consistent handling of documents of varying lengths.
Our window-size ablation study in \autoref{sec:ws-selection} supports this hypothesis.

Furthermore, this paper explores strategies to obtain high-quality attention weights and incorporates them into graph construction.
The key components of this process are: (1) \textbf{non-linear transformations} of the attention weights, and (2) \textbf{statistical filtering} to enforce sparsity.

Several previous research studies empirically demonstrated that activation functions can modulate raw attention weights, leading to more expressive results and capturing complex relationships between tokens.
They could introduce task-specific biases that benefit models in learning more than others \cite{sharma2017activation} (See \autoref{fig:attn}).

Similarly, the statistical filtering step helps to prune weak connections and retain the most salient dependencies.
\citeauthor{bugueno2025}’s results showed that mean-bound filtering showed optimal results for medium-length document processing, while max-bound filtering is optimal for long document processing.

\begin{figure*}[t]
  \includegraphics[width=\textwidth]{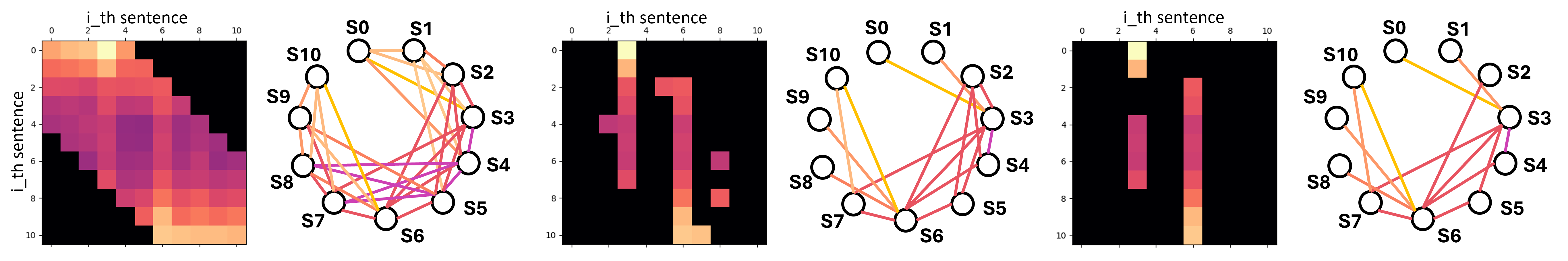}
  \caption{Attention matrix from a random document in the HND dataset paired with graphs constructed according to different statistical filtering: non-filtered (Left), mean-bound filter (Middle), and max-bound filter (Right). Brighter colors illustrate higher values.}
  \label{fig:graph_construction}
\end{figure*}

\subsection{Sliding Window Attention Models}

We first encode each document $D$ using a pretrained Sentence Transformer model, mapping every sentence $s_i$ to its corresponding embedding $x_i$. Then, the resulting set of embeddings $X$ constitutes the input to the SWA models.

The self-attention layer is implemented as a sliding-window multi-head attention mechanism that restricts attention computation to tokens within the defined window size or a predefined maximum cut-off length.
For the non-linear transformation, we evaluate multiple activation functions that have shown strong performance in previous studies (specifically ReLU \citep{wortsman2023replacing,bugueno2025} and Sigmoid \cite{ramapuram2024theory} in addition to the conventional Softmax, to assess their effect on model expressiveness and stability.

To compute scaled dot-product attention weights, we define each activation as follows:

\begin{equation}
\text{Attention}_{softmax} = \text{softmax}\left( \frac{A}{T} \right)
\end{equation}

\begin{equation}
\text{Attention}_{relu} = \frac{\text{ReLU}(A)}{d}
\end{equation} 

\begin{equation}
\text{Attention}_{sigmoid} = \sigma\left( A - \log d \right)
\end{equation} where \( A \) is the attention logits, \( T \) the temperature, and \( d = \text{dim}(A_{-1}) \) be the size of the last dimension of \( A \), corresponding to the document length measured in sentences.

The default temperature is set to 1.
However, since previous literature showed that annealing temperature helped gradually sharpen attention distributions during training \cite{caron2021emerging}, we also experimented with annealing scheduling, decreasing it by 0.0004 per batch, down to the minimum of 0.1, as seen in \autoref{annt}.

\begin{equation}
\label{annt}
T = \max\left(0.1,\ e^{-\lambda \cdot i} \right)
\end{equation} where \( \lambda \) is the annealing rate and \( i \) is the global iteration index.
Further implementation details on SWA models are provided in \autoref{sec:setup}.

\subsection{Graph Construction and Filtering}

A document graph is defined as $G = (V,E)$, where $V$ is the set of sentence nodes and $E$ is the set of edges. Nodes $v_i$ and $v_j$ are connected by an undirected weighted edge $\alpha_{ij}$ if the attention weight $W_{ij}$ exceeds a predefined threshold $\tau$ after being adjusted by the tolerance degree $\delta$.

We adopt the two thresholds--mean-bound and max-bound--from the previous work \cite{bugueno2025}.
Mean-bound filter sets a threshold slightly above a sentence's average attention score to retain more relevant dependencies.
On the other hand, max-bound sets a threshold slightly below a sentence's maximum attention score to retain only the strongest connections.
Each threshold is calculated row-wise, so that each node is connected to at least one node.
We also merge nodes with identical embeddings into a single node to reduce redundancy.
However, after all steps, if a node is only connected to itself, we divide the weight $\alpha_{ii}$ by 2, then allocate the weight to $\alpha_{i-1,i}$ and $\alpha_{i,i+1}$.

\autoref{fig:graph_construction} illustrates the graph construction process using each type of statistical filtering. 

\section{Experiments}

We hypothesize that an effective graph-based representation should encode sufficient task-relevant information to support downstream NLP tasks.
To evaluate this, we focus primarily on document classification, a common yet challenging task for assessing a model’s ability to capture document-level semantics and structural information.

We examine the proposed approach along three key aspects: (i) classification performance, (ii) graph structure, and (iii) sensitivity to window size.
Specifically, we compare graph variants generated by SWA models using different nonlinear activation functions, as well as graphs constructed with and without statistical filtering. We further analyze the resulting graph structures with a focus on computational efficiency.
To assess sensitivity to contextual scope, we conduct a window-size ablation on one dataset, varying the attention window from 10\% to 50\% of the document length, measured in sentences.
Additionally, we explore (iv) an extension of our approach to extractive document summarization to evaluate generalizability and identify task-specific limitations.

\begin{figure}[ht!]
\centering
  \includegraphics[width=0.9\linewidth]{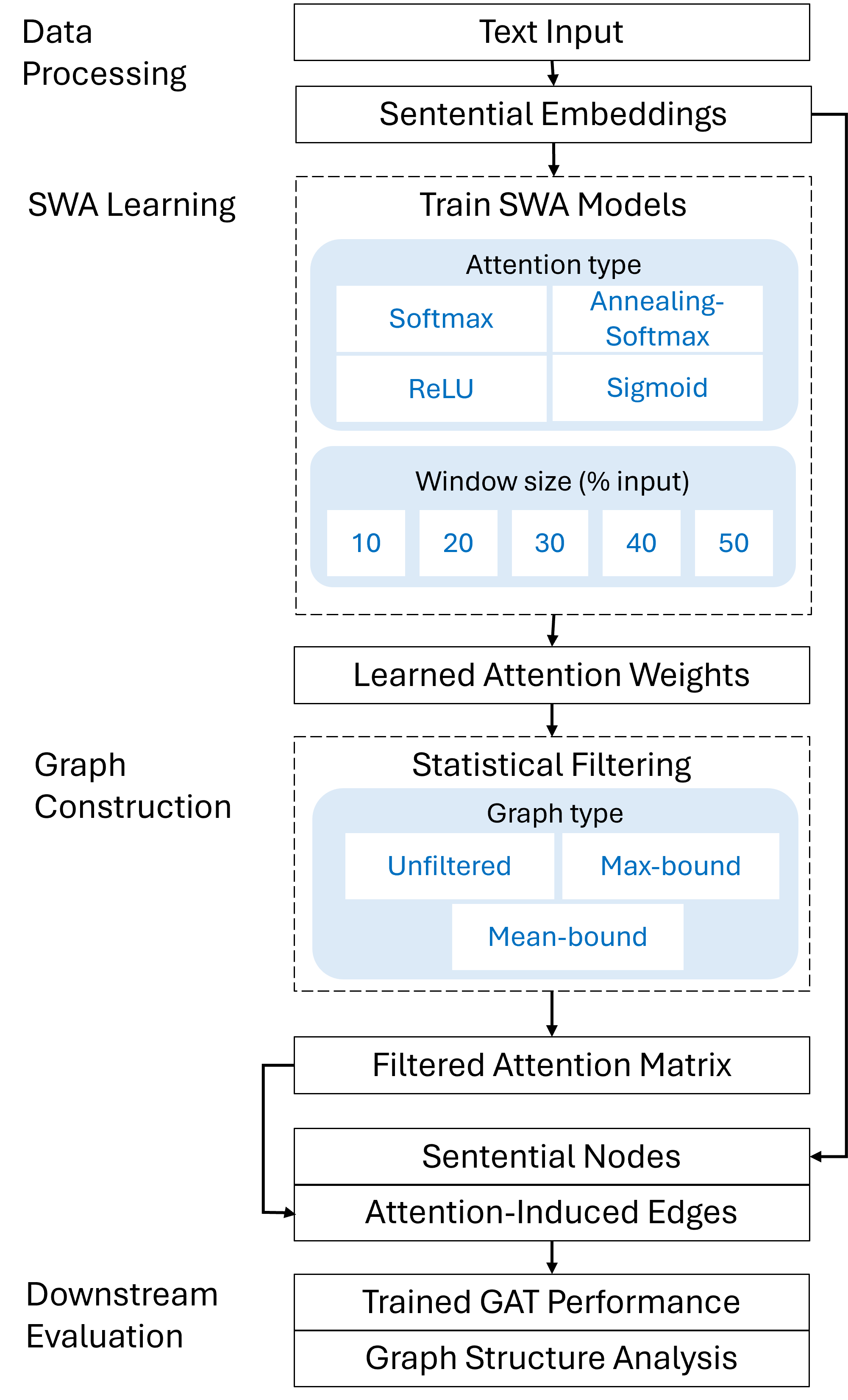}
  \caption{Overview of our experiment pipeline. The experimental components are illustrated in \textcolor{cyan}{blue}.}
  \label{fig:exp-overview}
\end{figure}

Across tasks, we train GAT models on our constructed graphs and evaluate performance using accuracy and F1 score, which respectively capture overall and class-level performance, particularly in class-imbalanced settings.
Implementation details of the GAT models are provided in \autoref{sec:setup}.
\autoref{fig:exp-overview} illustrates the proposed framework and the experiment pipeline.

\subsection{Datasets}
Three publicly available classification datasets are employed in our main experiments. To ensure direct comparability, we reuse all three document classification datasets from \citet{bugueno2025}.
A summary of the datasets is provided in \autoref{tab:datasets}, with further details in \autoref{appx:res_stat}.

\begin{itemize}[leftmargin=1.5em]
    \item \textbf{BBC News (BBC)} \cite{greene2006practical} comprises 2,225 medium-length news articles categorized into five relatively balanced topics: business, entertainment, politics, sports, and technology.
    \item \textbf{Hyperpartisan News Detection (HND)} \cite{kiesel2019semeval} contains 1,270 news articles with imbalanced labels of hyperpartisan or non-hyperpartisan texts.
    \item \textbf{arXiv classification (AX)}\footnote{Due to dataset size and resource constraints, we only experiment with Softmax attention-as the standard variant, and ReLU attention, to enable comparison with previous work.} \cite{he2019long} consists of 33,388 long scientific arXiv papers across 11 classes with slight imbalance.
\end{itemize}

\paragraph{Baselines} We use the heuristic baselines studied by \citet{bugueno2025}, including sentence-order, fixed-size window co-occurrence, mean-bound semantic similarity, and max-bound semantic similarity, along with their results.

\begin{table}
\small
\centering
\small
    \begin{tabular}{lrcrr}
    \hline
    \textbf{Dataset} & \textbf{K} & \textbf{Ratio} & \textbf{Length} & \textbf{Samples} \\
    \hline
    \ BBC & 5 & 4:5 & 19 & 2,225 \\
    \ HND & 2 & 1:2 & 21 & 1,270 \\
    \ AX & 11 & 1:2 & 539 & 33,388 \\ 
    \hline
    \end{tabular}
  \caption{Datasets summary. \textit{Ratio} denotes the class imbalance (major/minor), and \textit{Length} reports the average number of sentences per document.}
  \label{tab:datasets}
\end{table}

\section{Results} \label{results}

\autoref{sec:cls-performance} and \autoref{sec:graph-ana} present the results on document classification performance and graph efficiency, respectively, including direct comparisons with prior work \cite{bugueno2025}. 
\autoref{sec:ws-selection} details a window size ablation study on the HND dataset and its impact on model performance. 
Furthermore, \autoref{sec:res-summ} presents the results of exploratory experiments on document extractive summarization.
Additional results and corresponding analyses are provided in \autoref{sec:results}.

\begin{table*}[!htbp]
\centering
\small
\begin{tabularx}{\textwidth}{p{0.8cm}p{1.7cm}p{2.5cm}rrrrrrX}
\hline
\textbf{Dataset} & \textbf{Graph Type} & \textbf{Scheme} & \textbf{Acc} & \textbf{F1} & \textbf{|V|} & \textbf{|E|} &  \textbf{Degree} & \textbf{Disk} & \textbf{Variant} \\
\hline
\multirow{6}{*}{BBC} 
& heuristic & order & 96.5 & 96.3 & 19.4 & 36.6 & 1.9 & 74 M & 1-64 \\
& heuristic & window & 96.5 & 96.3 & 19.4 & 71.2 & 3.7 & 76 M & 1-64\\
& heuristic & mean semantic & 96.4 & 96.3 & 19.4 & 159.7 & 5.4 & 84 M & 1-256 \\
& heuristic & max semantic & 96.5 & 96.3 & 19.4 & 36.7 & 1.9 & 74 M & 1-64 \\
& full-attention & mean-bound \shortcite{bugueno2025} & 96.4 & 96.3 & 19.4 & 245.8 & 9.5 & 90 M & ReLU-3-128 \\
& full-attention & max-bound \shortcite{bugueno2025} & 96.3 & 96.1 & 19.4 & 66.3 & 3.4 & 77 M & ReLU-2-256\\
\cline{2-10}
& SW-attention & unfiltered & \textbf{96.8} & \textbf{96.7} & 19.4 & 540.8 & 27.8 & 109 M & Softmax-2-128\\
& SW-attention & mean-bound & 96.3 & 96.1 & 19.4 & 150.2 & 7.7 & 80 M & Anneal-3-64\\
& SW-attention & max-bound & 96.3 & 96.1 & 19.4 & 69.3 & 3.6 & 74 M & ReLU-3-128\\
\hline
\multirow{6}{*}{HND}
& heuristic & order & 76.3 & 76.3 & 20.3 & 37.0 & 1.8 & 43 M & 2-128 \\
& heuristic & window & 76.7 & 76.7 & 20.3 & 72.0 & 3.4 & 44 M & 2-256\\
& heuristic & mean semantic & 76.0 & 75.9 & 20.3 & 254.8 & 6.0 & 53 M & 2-256\\
& heuristic & max semantic & 75.9 & 75.9 & 20.3 & 36.9 & 1.8 & 43 M & 1-64\\
& full-attention & mean-bound \shortcite{bugueno2025} & 76.1 & 76.1 & 20.3 & 329.6 & 8.9 & 56 M & ReLU-2-256\\
& full-attention & max-bound \shortcite{bugueno2025} & 74.7 & 74.5 & 20.3 & 57.4 & 2.8 & 44 M & ReLU-2-128\\
\cline{2-10}
& SW-attention & unfiltered & \textbf{77.8} & \textbf{77.7} & 20.3 & 842.7 & 41.6 & 77 M & Sigmoid-2-64\\
& SW-attention & mean-bound & 77.0 & 77.0 & 20.3 & 222.8 & 11.0 & 51 M & Anneal-1-64\\
& SW-attention & max-bound & 75.8 & 75.8 & 20.3 & 66.2 & 3.3 & 44 M & Anneal-1-64\\
\hline
\multirow{4}{*}{AX}
& heuristic & order & 84.2 & 83.4 & 511.7 & 1035.0 & 2.0 & 25 GB & 3-64 \\
& heuristic & window & 83.7 & 83.0 & 511.7 & 1068.3 & 4.0 & 26 GB & 2-256\\
& heuristic & max semantic & 84.0 & 83.3 & 511.7 & 1241.9 & 2.3 & 26 GB & 3-64\\
& full-attention & max-bound \shortcite{bugueno2025} & \textbf{86.3} & \textbf{85.9} & 511.7 & 1092.2 & 2.2 & 25 GB & ReLU-3-64\\
\cline{2-10}
& SW-attention & mean-bound & 86.0 & 85.6 & 511.7 & 14662.3 & 28.7 & 40 GB & Softmax-2-64 \\
& SW-attention & max-bound & 84.6 & 84.2 & 511.7 & 1281.5 & 2.5 & 26 GB & Softmax-1-64 \\
\hline
\end{tabularx}
\caption{Average accuracy, F1 scores, and graph structural statistics on each dataset, compared to \citet{bugueno2025}. Variants consist of attention-type (if applicable), and the number of GAT layers and hidden units.}
\label{tab:main-res}
\end{table*}

\subsection{Classification Performance} \label{sec:cls-performance}

\paragraph{Baseline comparison.}
\autoref{tab:main-res} compares our \textit{best-}performing models against prior heuristic graphs baselines and learned graphs induced from full-attention models with statistical filtering.

On the medium-length BBC and HND datasets, graphs induced from sliding-window attention (SWA) weights outperform both baselines by 0.5-3 points on accuracy and F1 score.
Unfiltered SWA graphs yield the best results, indicating that restricting attention to local windows preserves nearly all task-relevant structural information for this medium-length, balanced dataset.
The large gains on HND suggest that learned local attention is particularly useful in more challenging settings, with class imbalance and stylistic variation rather than topic variation

A notable finding is that unfiltered SW-attention graphs perform best on both BBC and HND.
This contrasts with \citet{bugueno2025}, who found that statistically filtered full-attention graphs consistently improved classification performance.
This might suggest that, first, low-weight edges are not necessarily noise, since each node attends only to a limited local context.
Second, GAT models trained on such graphs may implicitly suppress irrelevant connections or exploit expressive yet non-salient dependencies that are subsequently removed by our statistical filters; for instance, salient dependencies that are eliminated under row-wise thresholding but would be preserved under a global criterion.

At the current scale and chosen window size, the increased graph density does not yet translate into prohibitive inefficiencies, although non-filtered graphs incur markedly higher memory and computational costs. 
These results exreveal a clear trade-off between computational cost and performance: non-filtered sliding-window attention graphs yield statistically significant gains (\texttt{p < 0.05}) at the expense of increased resource usage.

On AX, SWA-induced graphs substantially outperform the heuristic baselines, indicating that learned local dependencies are more informative than manually designed graph structures even for very long documents.
However, GAT performance on our learned graphs does not surpass the best full-attention-induced graphs, suggesting that certain long-range dependencies in scientific articles are difficult to recover using local window-based approaches alone. Nevertheless, the difference is relatively small, with only a 0.3 F1-score gap between the two strategies.

While full attention-induced graphs allow every sentence pair to interact and can therefore capture arbitrary long-range dependencies,
the number of candidate edges grows rapidly, leading to over-connectivity.
Therefore, statistical filtering becomes an essential component, by retaining only semantically salient edges, and recover a sparse backbone of robust dependencies. 

SWA-induced graphs, on the other hand, mitigate this issue by its locality constraint.
The limited number of candidate edges imposes an inductive bias that aligns with discourse structure, neighboring sentences often carry the strongest topical cues.
This results in denser, however, more efficient local neighborhoods without the combinatorial explosion of globally less informative connectivity. 

The main limitation of these locally-aware graphs is that it cannot directly model dependencies beyond the attention window.
Most apparently on the AX dataset, where scientific articles may contain semantically related sections separated by hundreds of sentences. 
Such cases suggest that this additional global context can provide a small but measurable advantage.

Overall, these results show that SW-attention graphs are a strong alternative to both heuristic graph construction and full-attention-based extraction. They deliver the best classification results on two datasets and remain competitive on the third, while relying on local attention patterns that are substantially more computationally efficient than unrestricted self-attention.

\paragraph{Model variants.}

To assess the effect of graph construction variants, we performed one-way ANOVA followed by Tukey's HSD tests over attention functions and statistical filters.
The detailed statistical analysis is elaborated in \autoref{appx:res_stat}.

The choice of activation function (Softmax, ReLU, Sigmoid, or Anneal) has limited practical impact.
\autoref{fig:attn} illustrates these differences qualitatively. 
Softmax attention tends to concentrate weights on a small number of highly connected nodes, creating hub-like structures. 
ReLU, Sigmoid, and Anneal produce more distributed attention patterns, yielding hierarchical connectivity with fewer dominant hubs. 
Although some pairwise differences are statistically significant, the overall performance variation within each dataset remains small. Despite these structural differences, downstream performance is largely comparable, indicating limited sensitivity to the specific choice of weight normalization, as long as the induced graph preserves a sufficient level of connectivity.

In contrast, filtering has a clearer effect. 
On BBC and HND, unfiltered graphs consistently outperform filtered variants, confirming that moderate-strength local edges contribute useful information. 
On AX, mean-bound filtering slightly improves over max-bound filtering and approaches the strongest overall result, indicating that some sparsification becomes beneficial as document length increases.
However, aggressive pruning does not provide uniform gains and can remove informative edges.

Across datasets, the variant analysis indicates that the proposed graph construction approach is robust across a broad range of configurations.
Statistical filtering matters more than the choice of attention function, but its effect depends on document length. Medium-length datasets benefit from preserving dense local context, whereas very long documents may require moderate sparsification to balance information retention and computational efficiency.

Overall, these findings suggest that future work should focus on adaptive sparsification strategies that preserve strong global dependencies while retaining informative low-weight local connections.
While our experiments use relatively simple GAT models trained from scratch, there are also potentials of further gains with more expressive graph architectures such as GraphSAGE \citep{hamilton2017inductive} or hierarchical graph encoders.

\subsection{Graph Structural Analysis} \label{sec:graph-ana}

We compare the graph structures induced by SWA models to \citet{bugueno2025}'s full attention-induced graphs.

\autoref{tab:main-res} shows that the strongest SWA-induced graphs generally correspond to graphs that retain substantially more edges than both heuristic baselines and filtered full-attention-induced graphs.
For the medium-length datasets, BBC and HND, unfiltered SW-attention produces dense local graphs with average degrees of 27.8 and 41.6, respectively, compared with degrees below 10 for all competing methods.
Despite this increase in connectivity, the storage overhead remains moderate. 
Thus, preserving dense local attention patterns is computationally practical.

On the same datasets, mean-bound SWA graphs obtain better GAT performance compared to the full-attention counterpart despite roughly 10\% smaller disk storage and 40\% fewer edges, implying less computational resource requirements.

Compared with full-attention-induced graphs, SW-attention graphs tend to distribute connectivity more evenly across nodes rather than concentrating edges around a few hubs.
This topology increases the number of alternative message-passing paths and reduces reliance on a few highly connected nodes.
In principle, a more balanced degree distribution should mitigate over-smoothing and over-squashing by reducing information bottlenecks and preserving multiple routes for contextual propagation.
This may help explain why SWA-induced graphs achieve strong performance with relatively shallow GAT architectures.

The AX dataset presents a different perspective.
In extremely long documents, graph density scales much more aggressively. 
While the mean-bound SWA graph requires much more storage, compared with the strongest full-attention baseline, 
the substantial increase in resource requirements brings diminishing returns: 
once the graph is sufficiently dense, additional edges become noisy and message-passing costs substantially.

Overall, the structural analysis indicates that SW-attention is most effective when it preserves a relatively rich connectivity pattern. 
The empirical results here show that the benefits outweigh the risks of redundant edges in medium-length documents. 
However, the computational cost of this strategy grows rapidly with document length.
The main practical challenge is therefore to balance structural richness against scalability.

\subsection{Window Size Ablation} \label{sec:ws-selection}

\begin{table}[]
\centering
\small
\begin{tabular}{c|c|ccc}
\hline
\multirow{2}{*}{\textbf{Window size}} & \multirow{2}{*}{\textbf{SWA}} & \multicolumn{3}{c}{\textbf{GAT}} \\
\cline{3-5}
 &  & unfilt & mean & max \\ \hline
10 & 72.5 & 76.2 & \textbf{75.6} & 76.0 \\
20 & \textbf{74.1} & 76.4 & 75.3 & 75.4 \\
30 & 73.4 & 76.2 & 75.1 & 73.6 \\
40 & 73.2 & 76.2 & 75.5 & 76.2 \\
50 & 73.7 & \textbf{76.5} & 75.3 & \textbf{76.6} \\
\hline
\end{tabular}
\caption{Accuracy scores on the \textbf{HND} dataset averaged over 5 independent runs, using the Softmax attention.
GAT results report the average of 3 graph types: unfiltered, mean-bound and max-bound.}
\label{tab:window-size}
\end{table}

We perform a window-size ablation to assess its effect in our pipeline. 
Since our model's performance on BBC is near optimal and AX is exceptionally large, we restrict this study to the HND dataset.

According to \autoref{tab:window-size}, GNN-based models consistently outperform the standalone SW-attention classifiers.
This confirms that converting attention patterns into graph structure provides a systematic benefit beyond using transformer representations alone (see also \autoref{tab:bbc-stat-ana}, \autoref{tab:hnd-stat-ana}, \autoref{tab:ax-stat-ana}).
Moreover, GAT classifiers are substantially more stable than the standalone SWA classifiers.

The effect of window size is modest but statistically significant (see \autoref{tab:ws-abl}).
Window sizes of 10 and 50 yield the strongest results and are statistically indistinguishable, whereas window sizes of 30 are consistently worse.
Because a window size of 30 is used in the main experiments, these findings indicate that the reported classification results should be interpreted as conservative estimates rather than an optimized hyperparameter.

Moderate window sizes do not necessarily produce better graphs than either smaller or larger ones.
These results highlight the complementary roles of local attention and graph message passing.
Although self-attention is restricted to a fixed context, the resulting graph allows information to propagate over multiple hops, effectively extending the receptive field beyond the original window.

The ablation study shows that window size influences absolute performance, but does not alter the broader conclusions of this work. 
In practice, window size should be treated as a tunable hyperparameter, but the effectiveness of the proposed graph construction approach does not depend on a narrowly optimized choice.

\subsection{Adaptation to Document Summarization} \label{sec:res-summ}

\begin{table*}[]
\centering
\small
\begin{tabular}{ll|ccccl}
\hline
\multirow{2}{*}{\textbf{Graph Type}} & \multirow{2}{*}{\textbf{Scheme}} & \multirow{2}{*}{\textbf{Acc}} & \multirow{2}{*}{\textbf{F1}} & \multicolumn{2}{c}{\textbf{F1 per class}} & \multirow{2}{*}{\textbf{Variant}} \\ 
\cline{5-6} & & & & \textbf{non-summary} & \textbf{summary} & \\ \hline
heuristic & order & 65.6 & 48.6 & 78.0 & 19.2 & 3-256 \\
heuristic & fixed window & \textbf{72.2} & \textbf{51.0} & \textbf{83.2} & 18.9 & 3-256 \\
\hline
SW-attention & mean-bound & 59.6 & 45.0 & 72.9 & 17.1 & Sigmoid-3-64 \\
SW-attention & max-bound & 63.0 & 47.7 & 75.9 & \textbf{19.5} & Softmax-2-256 \\ \hline
\end{tabular}
\caption{Performance comparison on the \textbf{GR} summarization task. F1 scores are reported at the macro and class levels. Variant consists of attention-type (if applicable), number of GAT layers, and number of hidden nodes.}
\label{tab:gr-results}
\end{table*}

To further assess the generality of the proposed graph construction method, we conduct an exploratory experiment on extractive document summarization, a task that also requires modeling document semantics and structure.

We conduct experiments on the GovReport (GR) dataset \cite{huang2021efficient}--a dataset that contains long U.S. Government reports--, as well as two heuristic baselines adopted from \citet{bugueno2025}: sentence-order graphs (\textit{order}) and fixed-size co-occurrence window graphs with window size 2 (\textit{fixed window}).
We frame summarization as a sentence classification task, predicting whether each sentence is summary-worthy (\texttt{class 1}) or not (\texttt{class 0}), and the selected sentences are then compared against oracle summaries \cite{liu2019text}, as described in \autoref{sec:gr-details}.

Because GovReport consists of substantially longer documents, we restrict our experiments to filtered graphs using mean-bound and max-bound sparsification.
This design choice reflects the computational constraints and further highlights the importance of scalable graph construction.

The results in \autoref{gr-gat} show that, unlike in document classification, the proposed attention-induced graphs do not outperform the simple order-based and fixed-window baselines.
The strongest graph-based configuration remains below both heuristic methods in terms of overall accuracy and F1 score.

Nevertheless, graph models remain competitive on the minority summary class.
Apparently, the induced graphs identify sentences that are semantically relevant to the reference summaries, even though it leads to over-predicting non-summary sentences and therefore lower overall accuracy (see \autoref{sec:gr-sentratio}).
In practice, the model often identifies informative sentences, but not necessarily the subset that best matches the oracle labels.

Several characteristics of the task help explain this result.
Many sentences not selected by the oracle remain partially relevant, making the distinction between positive and negative examples less clear than in document classification.
Second, the GAT is trained as an independent binary classifier for each sentence and does not explicitly model redundancy or summary length constraints.

This interpretation is supported by the representation analysis. 
A t-SNE projection of the learned sentence embeddings shows that summary sentences tend to occupy central regions of the embedding space rather than forming clearly separated clusters (see \autoref{fig:tsne-samples}). 
These sentences are semantically representative of the document content and therefore achieve strong semantic similarity scores such as BERTScore. 
However, semantically central sentences are not always the most informative under ROUGE-based evaluation, which favors concise and non-redundant coverage of key information (see \autoref{sec:gr-simscore}).

Better performance of multi-layered GATs benefits from multi-hop message passing, which allows each sentence representation to aggregate evidence from a broader portion of the document, introduced several sentences or sections earlier.
Therefore, extractive summarization depends more heavily on discourse position and section structure than on fine-grained semantic relations.
Important sentences frequently occur near introductions and conclusions, and such regularities can be captured effectively by simple order-based graphs. 
However, our analysis of sentence position distributions does not provide conclusive evidence that positional information alone explains the performance differences (see \autoref{sec:gr-sentdist}).

Although these results indicate that the proposed graph construction approach remains viable for extractive summarization, its benefits are currently more limited than in document classification.

Future work should focus less on refining edge construction and more on improving the sentence selection objective, for example through cost-sensitive or focal losses to address class imbalance, threshold calibration to match the oracle extraction ratio, ranking-based objectives, redundancy-aware decoding, and explicit incorporation of discourse and positional priors.

\section{Conclusion}

We present another data-driven approach for constructing graph-based document representations utilizing dynamic sliding-window attention models to automatically capture local and mid-range contextual relations between sentences.
We validate our approach by training GAT models on two common NLP tasks.
The results and following analyses demonstrate that, even with shallow attention models, 
the proposed strategy preserves structurally informative semantic dependencies while providing a computationally efficient alternative to existing graph construction methods.
These findings highlight the potential of automatically learned localized attention as a scalable basis for graph-based document modeling.

\section*{Limitations}

While our experiments demonstrate the effectiveness of the proposed approach, several aspects warrant further exploration. First, the current evaluation is limited to English-language datasets, primarily within the news domain and of medium to medium-long document length. Extending the analysis to additional languages and domains would provide a more comprehensive assessment of generalization.
Second, for the summarization task, evaluation was conducted against oracle summaries only, without direct comparison to human-written gold summaries or qualitative human assessments.
Adapting the approach to other downstream tasks and a deeper error analysis could also yield valuable insights into model behavior. Nonetheless, the observed results highlight the promise of our method and motivate future research toward broader applicability and more extensive evaluation.

\section*{Acknowledgments}
We sincerely thank the EACL 2026 Student Research Workshop (SRW) reviewers for their constructive feedback and insightful comments. We are also grateful to the EACL SRW organizing committee for creating a supportive and engaging environment for early-stage research and scholarly exchange.

\bibliography{custom}

@misc{bugueno2025,
      title={Rethinking Graph-Based Document Classification: Learning Data-Driven Structures Beyond Heuristic Approaches}, 
      author={Margarita Bugueño and Gerard de Melo},
      year={2025},
      eprint={2508.00864},
      archivePrefix={arXiv},
      primaryClass={cs.CL},
      url={https://arxiv.org/abs/2508.00864}, 
}

@article{bugueno2023connecting,
  title={Connecting the Dots: What Graph-Based Text Representations Work Best for Text Classification using Graph Neural Networks?},
  author={Bugue{\~n}o, Margarita and de Melo, Gerard},
  journal={arXiv preprint arXiv:2305.14578},
  year={2023}
}

@inproceedings{waswani2017attention,
  title={Attention is all you need},
  author={Waswani, A and Shazeer, N and Parmar, N and Uszkoreit, J and Jones, L and Gomez, A and Kaiser, L and Polosukhin, I},
  booktitle={NIPS},
  year={2017}
}

@article{zaheer2020big,
  title={Big bird: Transformers for longer sequences},
  author={Zaheer, Manzil and Guruganesh, Guru and Dubey, Kumar Avinava and Ainslie, Joshua and Alberti, Chris and Ontanon, Santiago and Pham, Philip and Ravula, Anirudh and Wang, Qifan and Yang, Li and others},
  journal={Advances in neural information processing systems},
  volume={33},
  pages={17283--17297},
  year={2020}
}

@article{beltagy2020longformer,
  title={Longformer: The long-document transformer},
  author={Beltagy, Iz and Peters, Matthew E and Cohan, Arman},
  journal={arXiv preprint arXiv:2004.05150},
  year={2020}
}

@article{ma2022multi,
  title={Multi-document summarization via deep learning techniques: A survey},
  author={Ma, Congbo and Zhang, Wei Emma and Guo, Mingyu and Wang, Hu and Sheng, Quan Z},
  journal={ACM Computing Surveys},
  volume={55},
  number={5},
  pages={1--37},
  year={2022},
  publisher={ACM New York, NY}
}

@article{wang2024graph,
  title={Graph neural networks for text classification: A survey},
  author={Wang, Kunze and Ding, Yihao and Han, Soyeon Caren},
  journal={Artificial intelligence review},
  volume={57},
  number={8},
  pages={190},
  year={2024},
  publisher={Springer}
}

@article{hudson2022muld,
  title={Muld: The multitask long document benchmark},
  author={Hudson, G Thomas and Moubayed, Noura Al},
  journal={arXiv preprint arXiv:2202.07362},
  year={2022}
}

@article{van2016wavenet,
  title={Wavenet: A generative model for raw audio},
  author={Van Den Oord, Aaron and Dieleman, Sander and Zen, Heiga and Simonyan, Karen and Vinyals, Oriol and Graves, Alex and Kalchbrenner, Nal and Senior, Andrew and Kavukcuoglu, Koray and others},
  journal={arXiv preprint arXiv:1609.03499},
  volume={12},
  pages={1},
  year={2016}
}

@article{kovaleva2019revealing,
  title={Revealing the dark secrets of BERT},
  author={Kovaleva, Olga and Romanov, Alexey and Rogers, Anna and Rumshisky, Anna},
  journal={arXiv preprint arXiv:1908.08593},
  year={2019}
}

@article{xu2021contrastive,
  title={Contrastive document representation learning with graph attention networks},
  author={Xu, Peng and Chen, Xinchi and Ma, Xiaofei and Huang, Zhiheng and Xiang, Bing},
  journal={arXiv preprint arXiv:2110.10778},
  year={2021}
}

@inproceedings{wang2023docgraphlm,
  title={Docgraphlm: Documental graph language model for information extraction},
  author={Wang, Dongsheng and Ma, Zhiqiang and Nourbakhsh, Armineh and Gu, Kang and Shah, Sameena},
  booktitle={Proceedings of the 46th International ACM SIGIR Conference on Research and Development in Information Retrieval},
  pages={1944--1948},
  year={2023}
}

@inproceedings{mihalcea2004textrank,
  title={Textrank: Bringing order into text},
  author={Mihalcea, Rada and Tarau, Paul},
  booktitle={Proceedings of the 2004 conference on empirical methods in natural language processing},
  pages={404--411},
  year={2004}
}

@article{hassan2007random,
  title={Random walk term weighting for improved text classification},
  author={Hassan, Samer and Mihalcea, Rada and Banea, Carmen},
  journal={International Journal of Semantic Computing},
  volume={1},
  number={04},
  pages={421--439},
  year={2007},
  publisher={World Scientific}
}

@article{erkan2004lexrank,
  title={Lexrank: Graph-based lexical centrality as salience in text summarization},
  author={Erkan, G{\"u}nes and Radev, Dragomir R},
  journal={Journal of artificial intelligence research},
  volume={22},
  pages={457--479},
  year={2004}
}

@article{castillo2017text,
  title={Text analysis using different graph-based representations},
  author={Castillo, Esteban and Cervantes, Ofelia and Vilarino, Darnes},
  journal={Computaci{\'o}n y Sistemas},
  volume={21},
  number={4},
  pages={581--599},
  year={2017},
  publisher={Instituto Polit{\'e}cnico Nacional, Centro de Investigaci{\'o}n en Computaci{\'o}n}
}

@article{plenz2024graph,
  title={Graph Language Models},
  author={Plenz, Moritz and Frank, Anette},
  journal={arXiv preprint arXiv:2401.07105},
  year={2024}
}

@article{zhang2020every,
  title={Every document owns its structure: Inductive text classification via graph neural networks},
  author={Zhang, Yufeng and Yu, Xueli and Cui, Zeyu and Wu, Shu and Wen, Zhongzhen and Wang, Liang},
  journal={arXiv preprint arXiv:2004.13826},
  year={2020}
}

@article{onan2023hierarchical,
  title={Hierarchical graph-based text classification framework with contextual node embedding and BERT-based dynamic fusion},
  author={Onan, Aytu{\u{g}}},
  journal={Journal of king saud university-computer and information sciences},
  volume={35},
  number={7},
  pages={101610},
  year={2023},
  publisher={Elsevier}
}

@inproceedings{huang2022contexting,
  title={ConTextING: granting document-wise contextual embeddings to graph neural networks for inductive text classification},
  author={Huang, Yen-Hao and Chen, Yi-Hsin and Chen, Yi-Shin},
  booktitle={Proceedings of the 29th international conference on computational linguistics},
  pages={1163--1168},
  year={2022}
}

@article{velivckovic2017graph,
  title={Graph attention networks},
  author={Veli{\v{c}}kovi{\'c}, Petar and Cucurull, Guillem and Casanova, Arantxa and Romero, Adriana and Lio, Pietro and Bengio, Yoshua},
  journal={arXiv preprint arXiv:1710.10903},
  year={2017}
}

@article{qian2018graphie,
  title={Graphie: A graph-based framework for information extraction},
  author={Qian, Yujie and Santus, Enrico and Jin, Zhijing and Guo, Jiang and Barzilay, Regina},
  journal={arXiv preprint arXiv:1810.13083},
  year={2018}
}

@inproceedings{yao2019graph,
  title={Graph convolutional networks for text classification},
  author={Yao, Liang and Mao, Chengsheng and Luo, Yuan},
  booktitle={Proceedings of the AAAI conference on artificial intelligence},
  volume={33},
  number={01},
  pages={7370--7377},
  year={2019}
}

@article{zeng2020double,
  title={Double graph based reasoning for document-level relation extraction},
  author={Zeng, Shuang and Xu, Runxin and Chang, Baobao and Li, Lei},
  journal={arXiv preprint arXiv:2009.13752},
  year={2020}
}

@inproceedings{hua2023improving,
  title={Improving long dialogue summarization with semantic graph representation},
  author={Hua, Yilun and Deng, Zhaoyuan and McKeown, Kathleen},
  booktitle={Findings of the Association for Computational Linguistics: ACL 2023},
  pages={13851--13883},
  year={2023}
}

@article{sharma2017activation,
  title={Activation functions in neural networks},
  author={Sharma, Sagar and Sharma, Simone and Athaiya, Anidhya},
  journal={Towards Data Sci},
  volume={6},
  number={12},
  pages={310--316},
  year={2017}
}

@inproceedings{kiesel2019semeval,
  title={SemEval-2019 task 4: Hyperpartisan news detection},
  author={Kiesel, Johannes and Mestre, Maria and Shukla, Rishabh and Vincent, Emmanuel and Adineh, Payam and Corney, David and Stein, Benno and Potthast, Martin},
  booktitle={Proceedings of the 13th International Workshop on Semantic Evaluation},
  pages={829--839},
  year={2019}
}

@article{huang2021efficient,
  title={Efficient attentions for long document summarization},
  author={Huang, Luyang and Cao, Shuyang and Parulian, Nikolaus and Ji, Heng and Wang, Lu},
  journal={arXiv preprint arXiv:2104.02112},
  year={2021}
}

@article{liu2019text,
  title={Text summarization with pretrained encoders},
  author={Liu, Yang and Lapata, Mirella},
  journal={arXiv preprint arXiv:1908.08345},
  year={2019}
}

@inproceedings{greene2006practical,
  title={Practical solutions to the problem of diagonal dominance in kernel document clustering},
  author={Greene, Derek and Cunningham, P{\'a}draig},
  booktitle={Proceedings of the 23rd international conference on Machine learning},
  pages={377--384},
  year={2006}
}

@article{kingma2014adam,
  title={Adam: A method for stochastic optimization},
  author={Kingma, Diederik P and Ba, Jimmy},
  journal={arXiv preprint arXiv:1412.6980},
  year={2014}
}

@inproceedings{lin2004rouge,
  title={Rouge: A package for automatic evaluation of summaries},
  author={Lin, Chin-Yew},
  booktitle={Text summarization branches out},
  pages={74--81},
  year={2004}
}

@article{zhang2019bertscore,
  title={Bertscore: Evaluating text generation with bert},
  author={Zhang, Tianyi and Kishore, Varsha and Wu, Felix and Weinberger, Kilian Q and Artzi, Yoav},
  journal={arXiv preprint arXiv:1904.09675},
  year={2019}
}

@article{wortsman2023replacing,
  title={Replacing softmax with relu in vision transformers},
  author={Wortsman, Mitchell and Lee, Jaehoon and Gilmer, Justin and Kornblith, Simon},
  journal={arXiv preprint arXiv:2309.08586},
  year={2023}
}

@article{ramapuram2024theory,
  title={Theory, analysis, and best practices for sigmoid self-attention},
  author={Ramapuram, Jason and Danieli, Federico and Dhekane, Eeshan and Weers, Floris and Busbridge, Dan and Ablin, Pierre and Likhomanenko, Tatiana and Digani, Jagrit and Gu, Zijin and Shidani, Amitis and others},
  journal={arXiv preprint arXiv:2409.04431},
  year={2024}
}

@inproceedings{caron2021emerging,
  title={Emerging properties in self-supervised vision transformers},
  author={Caron, Mathilde and Touvron, Hugo and Misra, Ishan and J{\'e}gou, Herv{\'e} and Mairal, Julien and Bojanowski, Piotr and Joulin, Armand},
  booktitle={Proceedings of the IEEE/CVF international conference on computer vision},
  pages={9650--9660},
  year={2021}
}

@inproceedings{bugueno2025graphlss,
  title={GraphLSS: Integrating Lexical, Structural, and Semantic Features for Long Document Extractive Summarization},
  author={Bugue{\~n}o, Margarita and Abou Hamdan, Hazem and De Melo, Gerard},
  booktitle={Proceedings of the 2025 Conference of the Nations of the Americas Chapter of the Association for Computational Linguistics: Human Language Technologies (Volume 2: Short Papers)},
  pages={797--804},
  year={2025}
}

@article{he2019long,
  title={Long document classification from local word glimpses via recurrent attention learning},
  author={He, Jun and Wang, Liqun and Liu, Liu and Feng, Jiao and Wu, Hao},
  journal={IEEE Access},
  volume={7},
  pages={40707--40718},
  year={2019},
  publisher={IEEE}
}

@article{buterez2025end,
  title={An end-to-end attention-based approach for learning on graphs},
  author={Buterez, David and Janet, Jon Paul and Oglic, Dino and Li{\`o}, Pietro},
  journal={Nature Communications},
  volume={16},
  number={1},
  pages={5244},
  year={2025},
  publisher={Nature Publishing Group UK London}
}

@article{hamilton2017inductive,
  title={Inductive representation learning on large graphs},
  author={Hamilton, Will and Ying, Zhitao and Leskovec, Jure},
  journal={Advances in neural information processing systems},
  volume={30},
  year={2017}
}

@incollection{fisher1970statistical,
  title={Statistical methods for research workers},
  author={Fisher, Ronald Aylmer},
  booktitle={Breakthroughs in statistics: Methodology and distribution},
  pages={66--70},
  year={1970},
  publisher={Springer}
}

@article{tukey1949comparing,
  title={Comparing individual means in the analysis of variance},
  author={Tukey, John W},
  journal={Biometrics},
  pages={99--114},
  year={1949},
  publisher={JSTOR}
}

\appendix

\section{Datasets}
\label{appx:data}

\subsection{Document Classification Datasets}

\textbf{BBC News} \cite{greene2006practical} is a collection of 2,225 news articles from the BBC News website. The articles are considered medium-length and are labeled according to the five topics: business, entertainment, politics, sport, and technology. The labels are relatively balanced. 

\textbf{Hyperpartisan News Detection} (HND) \cite{kiesel2019semeval} is a collection of 645 news articles, with binary labels as hyperpartisan or non-hyperpartisan, categorizing if a news article ignores or attacks the opposing views blindly. The dataset is relatively small and imbalanced.
Moreover, the dataset contains some relatively long documents, namely, over 5\% of the documents contain more than 50 sentences.

Extended \textbf{arXiv document classification (AX)} \cite{he2019long}  consisting of 33,388 papers across 11 subject categories in Mathematics and Computer Science.
The label expands a previously released 4-class version to a more diverse and challenging multi-class setting through a weakly supervised method.
Each document contains at least 1,000 words and is truncated within 10,000 words.

\subsection{Document Summarization Dataset}
\label{sec:gr-details}
\textbf{GovReport} (hence, GR) \cite{huang2021efficient} is a collection of 19,466 reports from the U.S. Government Accountability Office, covering a wide range of national policy issues.
The gold summaries were human-written and abstractive.

To adapt to the current extractive summarization task, oracle summaries \cite{liu2019text} have been prepared by including sentences that maximize the Rouge-1 score \cite{lin2004rouge}; in other words, the overlap between selected sentences and the original summaries.

We then frame the summarization task as a sentence classification task, where each sentence in a document is individually labeled as either to be included in the summary or not.
While a report contains an average of 300 sentences, approximately 10 are included in the summary. Therefore, the corresponding extractive sentence labels are extremely unbalanced. 

\section{Implementation Details}
\label{sec:setup}
The project is implemented mainly in \texttt{PyTorch} framework.
\texttt{PyTorch Geometric} is used to construct graph datasets and train GAT models.

The experiments on BBC and HND datasets were run on Google Colab T4. The experiments on GR and AX were run on NVIDIA GeForce RTX 3090.

\subsection{Setup}

Learned attention weights used to construct attention-induced graphs are extracted from multi-head SWA models. 
These graphs are then used to train Graph Attention Network (GAT) models in order to evaluate the efficiency of our approach. 
We deliberately employ a simple SWA architecture to demonstrate the effectiveness of the data-driven method at a foundational level, serving as a baseline for more advanced extensions.
In parallel, we evaluate multiple GAT architectures to assess the robustness of the proposed graph construction strategy across different model configurations.

Nonetheless, both types of models, the learning rate is 0.001. The batch size is set to 32.

Dropout is applied to each fully connected layer with a rate of 0.2.
Class weights are applied to the cross-entropy loss function to reduce bias toward the majority classes.

\subsection{Multi-Head SWA Models}
Sentence embeddings are obtained using a frozen pretrained model: \texttt{SentenceTransformer} (\texttt{paraphrase-MiniLM-L6-v2})\footnote{\url{https://huggingface.co/sentence-transformers/paraphrase-MiniLM-L6-v2}} with its 384 embedding dimensions.
A multi-head self-attention (MHA) model is comprised of the following trainable layers: two fully connected layers of 128 hidden nodes with Xavier-uniform initialization, a sliding-window attention layer, and a non-linear layer.
We employ 4 self-attention heads, and compute attention using scaled dot-product; then, we concatenate and pass it through the output projection layer.

The model can be trained up to 20 epochs using Adam optimization \cite{kingma2014adam}; however, if the macro-averaged F1 score on the validation split does not improve by at least 0.001 during the following five epochs, the training is interrupted.

For each setting, we initialize and train five models independently and evaluate them against the F1 score on the validation set.
Later, the learned attention weights from the MHA models with consistently higher F1 scores are used to construct graphs.

\subsection{Graph Construction and GAT models}

To prevent node isolation in the filtered graphs, we compute row-wise mean-bound and max-bound thresholds and apply a tolerance degree of 0.5 in all graph construction settings.
For the AX dataset, however, the tolerance degree for the mean-bound graph is increased to 1.0 to improve resource efficiency.

We experiment with the number of GAT layers and hidden nodes, from \texttt{\{1, 2, 3\}} and \texttt{\{64, 128, 256\}} respectively.
The models can be trained up to 50 epochs, but they are interrupted if the F1 score on the validation set does not improve by at least 0.001 in the five consecutive epochs.
For each GAT setting, we obtain results from five independent runs and average them.

\section{Detailed GAT Results and Statistical Analysis}
\label{appx:res_stat}

Here we report average GAT results with the two studied components, i.e., attention types and statistical filters, in \autoref{bbc-gat} (BBC), \autoref{hnd-gat} (HND), \autoref{ax-gat} (AX), \autoref{wshnd-gat} (window ablation) and \autoref{gr-gat} (summarization).
However, our study explores various combinations of non-linear activation functions, hidden layers, hidden nodes, and statistical filters, with 5 independent runs per configuration\footnote{All detailed results are kept in the \texttt{GNN\_Results} sub-folder in our GitHub repository.}.

In addition, we conduct one-way ANOVA \citep{fisher1970statistical} followed by post-hoc Tukey's honestly significant difference (HSD) test \citep{tukey1949comparing}\footnote{Implemented using Python's \texttt{scipy.stats} library: \url{https://docs.scipy.org/doc/scipy/reference/stats.html}.} to assess whether differences among variants are statistically significant for the components of interest, namely activation functions, statistical filters, and their combinations.
All findings are reported in \autoref{tab:bbc-stat-ana} (BBC), \autoref{tab:hnd-stat-ana} (HND), \autoref{tab:ax-stat-ana} (AX), and \autoref{tab:gr-stat-ana} (GR).

\paragraph{BBC}

GAT models consistently outperform SWA models, indicating gains from our graph construction approach.

Performance differences across attention activation functions are negligible. Anneal, Softmax, ReLU, and Sigmoid yield nearly identical results.
Thus, the specific non-linear transformation applied to the raw attention scores has little practical effect once the graph is constructed.

The filtering strategy, however, has a consistent and statistically significant effect on both accuracy and macro-F1 ($p < 0.001$). 
Across all attention types, unfiltered graphs achieve the strongest results, while both mean-bound and max-bound filtering lead to small reductions in performance.
Post-hoc comparisons confirm that unfiltered graphs significantly outperform both filtered variants, whereas the difference between mean-bound and max-bound filtering is rather limited.

The \textit{sport} class remains near ceiling across all configurations.
The more challenging \textit{business} and \textit{politics} categories suggest that weaker attention edges provide useful contextual information when class boundaries are less distinct.

\paragraph{HND}

No single attention function consistently performs best across all filtering conditions.
Anneal achieves the strongest result under mean-bound filtering, Softmax is more sensitive to aggressive pruning, ReLU gain significantly largest from their SWA classifiers, and Sigmoid yields the most stable performance, 
Post-hoc analysis, however, shows that ReLU- and Sigmoid-based attention consistently outperform Softmax, suggesting that non-normalized attention transformations preserve stronger contrast between salient and non-salient relationships.

The filtering strategy has a slightly larger practical effect than the choice of attention activation.
Mean-bound filtering produces the best overall configuration, indicating that moderate sparsification can improve the signal-to-noise ratio by removing weak or redundant edges while preserving informative semantic relationships.

\paragraph{AX}

GAT models significantly outperform SWA models in 3 out of 4 settings.
Both attention activation and filtering have no statistically significant effects.

Per-class performance is largely stable across configurations, with the exception of Computer Vision, which remains the most difficult regardless of the graph construction strategy.
This result suggests that classification difficulty is driven primarily by semantic overlap between scientific domains rather than by the specific graph variant.

\paragraph{Window Ablation (HND)}

Graph construction provides the largest benefit for GAT models when combined with smaller receptive fields, compared to SWA models trained on sentence embeddings.
However, the effect size decreases as the receptive field grows, and the improvement is not consistently maintained for larger windows, i.e., 50\%.

Window sizes 10, 20, 40, and 50 form a statistically indistinguishable group, with only minor fluctuations in both accuracy and F1 score.
This result suggests that the moderate windows do not necessarily provide the best balance between local and global context.

Unfiltered graphs and mean-bound graphs vary similarly across window sizes.
Contrastingly, max-bound filtering exhibits the largest performance variance. 
This indicates that aggressive pruning in the underlying attention distribution can lead to larger structural differences when only the strongest edges are retained.

\begin{table}
\centering
\small
\begin{tabular}{c|ccccc}
\hline
WS & 10 & 20 & 30 & 40 & 50    \\ \hline
10 & & 0.500 & \textbf{\textless{}0.001} & 0.894 & 0.425 \\
20 & 0.430 & & \textbf{\textless{}0.001} & 0.633 & 0.898 \\
30 & \textbf{\textless{}0.001} & \textbf{\textless{}0.001} & & \textbf{\textless{}0.001} & \textbf{\textless{}0.001} \\
40 & 0.833 & 0.611 & \textbf{\textless{}0.001} & & 0.554 \\
50 & 0.349 & 0.874 & \textbf{\textless{}0.001} & 0.516 & \\  \hline   
\end{tabular}
\caption{T-test results of GAT models' performance when training on \textbf{HND} graphs with different window sizes (WS). The bottom and left side measure the accuracy scores, and the top and right side measure the F1 scores on the test set. Statistical significance is highlighted in \textbf{bold}.}
\label{tab:ws-abl}
\end{table}

\paragraph{Extractive Summarization (GR)}

The relative behavior of attention functions remains stable across model depths and hidden dimensions.
The only statistically detectable main effect of attention funtions is a modest macro-F1 advantage of ReLU over Sigmoid ($p < 0.05$), while no significant differences are observed for accuracy.

In contrast, max-bound filtering significantly outperforms mean-bound filtering for both accuracy and macro-F1 ($p < 0.01$).
This pattern holds for every attention function and is highly consistent, indicating that more selective pruning yields slightly more effective sentence graphs for this task.

\begin{table*}[]
\centering
\small
\resizebox{\textwidth}{!}{%
\begin{tabular}{ll|cccccccccc}
\hline
\multicolumn{1}{c}{\multirow{2}{*}{\textbf{Attention}}} & \multicolumn{1}{c|}{\multirow{2}{*}{\textbf{Filter}}} & \multicolumn{1}{c}{\multirow{2}{*}{\textbf{Acc}}} & \multicolumn{1}{c}{\multirow{2}{*}{\textbf{Acc (std)}}} & \multicolumn{1}{c}{\multirow{2}{*}{\textbf{F1}}} & \multicolumn{1}{c}{\multirow{2}{*}{\textbf{F1 (std)}}} & \multicolumn{5}{c}{\textbf{F1 per class}} \\ \cline{7-11} 
\multicolumn{1}{c}{} & \multicolumn{1}{c|}{} & \multicolumn{1}{c}{} & \multicolumn{1}{c}{} & \multicolumn{1}{c}{} & \multicolumn{1}{c}{} & \textbf{sport} & \textbf{entertain} & \textbf{business} & \textbf{tech} & \textbf{politics} \\ \hline
\multirow{3}{*}{\textbf{Anneal}} & \textbf{unfilt} & 96.32**  & 0.66  & 96.14  & 0.70  & 99.01  & 96.15  & 94.67  & 96.34  & 94.55 \\
 & \textbf{max} & 95.84**  & 0.55  & 95.72  & 0.60  & 97.90  & 96.30  & 93.88  & 96.17  & 94.36 \\
 & \textbf{mean} & 95.87**  & 0.39  & 95.73  & 0.42  & 97.94  & 95.95  & 94.52  & 96.38  & 93.86 \\ \hline
 \multirow{3}{*}{\textbf{Softmax}} & \textbf{unfilt} & 96.32**  & 0.55  & 96.14  & 0.57  & 99.02  & 96.27  & 94.60  & 96.27  & 94.53 \\
 & \textbf{max} & 95.73**  & 0.60  & 95.61  & 0.65  & 97.81  & 96.20  & 93.80  & 95.79  & 94.44 \\
 & \textbf{mean} & 95.68**  & 0.47  & 95.51  & 0.49  & 98.08  & 95.53  & 94.16  & 96.00  & 93.76 \\ \hline
\multirow{3}{*}{\textbf{ReLU}} & \textbf{unfilt} & 96.30**  & 0.60  & 96.11  & 0.64  & 99.08  & 96.08  & 94.62  & 96.13  & 94.64 \\
 & \textbf{max} & 95.89**  & 0.48  & 95.75  & 0.49  & 98.17  & 96.01  & 94.21  & 96.14  & 94.19 \\
 & \textbf{mean} & 95.46**  & 0.72  & 95.24  & 0.76  & 98.22  & 94.57  & 94.28  & 95.62  & 93.48 \\ \hline
\multirow{3}{*}{\textbf{Sigmoid}} & \textbf{unfilt} & 96.45**  & 0.56  & 96.28  & 0.59  & 99.05  & 96.32  & 94.74  & 96.38  & 94.93 \\
 & \textbf{max} & 95.69**  & 0.62  & 95.59  & 0.65  & 97.83  & 96.49  & 93.32  & 96.64  & 93.68 \\
 & \textbf{mean} & 95.71**  & 0.52  & 95.53  & 0.54  & 98.22  & 95.42  & 94.24  & 96.21  & 93.57 \\ \hline
\end{tabular}
}
\caption{GAT results of \textbf{BBC} dataset, with each attention type and filters, averaged across 5 independent runs, numbers of layers (\{1, 2, 3\}) and hidden nodes (\{64, 128, 256\}). * indicates a statistically significant improvement of the GAT accuracy over the SWA models ($p < 0.05$). ** indicate higher improvement ($p < 0.01$).}
\label{bbc-gat}
\end{table*}

\begin{table*}[]
\centering
\small
\begin{tabular}{ll|cccccc}
\hline
\multicolumn{1}{c}{\multirow{2}{*}{\textbf{Attention}}} & \multicolumn{1}{c|}{\multirow{2}{*}{\textbf{Filter}}} & \multicolumn{1}{c}{\multirow{2}{*}{\textbf{Acc}}} & \multicolumn{1}{c}{\multirow{2}{*}{\textbf{Acc (std)}}} & \multicolumn{1}{c}{\multirow{2}{*}{\textbf{F1}}} & \multicolumn{1}{c}{\multirow{2}{*}{\textbf{F1 (std)}}} & \multicolumn{2}{c}{\textbf{F1 per class}} \\ \cline{7-8} 
\multicolumn{1}{c}{} & \multicolumn{1}{c|}{} & \multicolumn{1}{c}{} & \multicolumn{1}{c}{} & \multicolumn{1}{c}{} & \multicolumn{1}{c}{} & \textbf{non-hyperpartisan} & \textbf{hyperpartisan} \\ \hline
\multirow{3}{*}{\textbf{Anneal}} & \textbf{unfilt} & 75.69  & 1.97  & 75.57  & 2.06  & 75.65  & 75.49 \\
 & \textbf{max} & 76.05  & 1.76  & 75.97  & 1.80  & 75.98  & 75.96 \\
 & \textbf{mean} & 77.00  & 1.72  & 76.93  & 1.78  & 76.83  & 77.03 \\ \hline
\multirow{3}{*}{\textbf{Softmax}} & \textbf{unfilt} & 76.25**  & 1.77  & 76.17  & 1.80  & 76.18  & 76.17 \\
 & \textbf{max} & 73.64  & 2.04  & 73.53  & 2.12  & 73.63  & 73.43 \\
 & \textbf{mean} & 75.10*  & 1.38  & 75.01  & 1.41  & 75.07  & 74.95 \\ \hline
\multirow{3}{*}{\textbf{ReLU}} & \textbf{unfilt} & 76.24**  & 1.32  & 76.14  & 1.35  & 75.90  & 76.38 \\
 & \textbf{max} & 75.90**  & 1.95  & 75.82  & 1.98  & 74.84  & 76.80 \\
 & \textbf{mean} & 76.00**  & 1.57  & 75.88  & 1.66  & 75.55  & 76.22 \\ \hline
\multirow{3}{*}{\textbf{Sigmoid}} & \textbf{unfilt} & 76.04  & 1.65  & 75.95  & 1.73  & 76.31  & 75.59 \\
 & \textbf{max} & 76.29  & 1.51  & 76.23  & 1.52  & 75.79  & 76.67 \\
 & \textbf{mean} & 76.08  & 1.38  & 76.00  & 1.47  & 75.84  & 76.16 \\ \hline
\end{tabular}
\caption{GAT results of \textbf{HND} dataset, with each attention type and filters, averaged across 5 independent runs, numbers of layers (\{1, 2, 3\}) and hidden nodes (\{128, 256\}). * indicates a statistically significant improvement of the GAT accuracy over the SWA models ($p < 0.05$).  ** indicate higher improvement ($p < 0.01$).}
\label{hnd-gat}
\end{table*}

\begin{table*}[]
\centering
\small
\resizebox{\textwidth}{!}{%
\begin{tabular}{ll|ccccccccccccccc}
\hline
\multicolumn{1}{c}{\multirow{2}{*}{\textbf{Attention}}} & \multicolumn{1}{c|}{\multirow{2}{*}{\textbf{Filter}}} & \multicolumn{1}{c}{\multirow{2}{*}{\textbf{Acc}}} & \multicolumn{1}{c}{\multirow{2}{*}{\textbf{Acc (std)}}} & \multicolumn{1}{c}{\multirow{2}{*}{\textbf{F1}}} & \multicolumn{1}{c}{\multirow{2}{*}{\textbf{F1 (std)}}} & \multicolumn{11}{c}{\textbf{F1 per class}} \\
\multicolumn{1}{c}{} & \multicolumn{1}{c|}{} & \multicolumn{1}{c}{} & \multicolumn{1}{c}{} & \multicolumn{1}{c}{} & \multicolumn{1}{c}{} & \textbf{0} & \textbf{1} & \textbf{2} & \textbf{3} & \textbf{4} & \textbf{5} & \textbf{6} & \textbf{7} & \textbf{8} & \textbf{9} & \textbf{10} \\ \hline
\multirow{2}{*}{\textbf{Softmax}} & \textbf{max} & 85.50**  & 0.76  & 85.06  & 0.74  & 94.21  & 84.31  & 65.64  & 87.17  & 94.66  & 82.63  & 89.42  & 86.29  & 88.26  & 74.68  & 88.38 \\
& \textbf{mean} & 85.27  & 0.67  & 84.86  & 0.71  & 94.33  & 84.58  & 65.00  & 86.98  & 95.24  & 81.88  & 89.35  & 85.63  & 87.17  & 75.11  & 88.19 \\ \hline
\multirow{2}{*}{\textbf{ReLU}} & \textbf{max} & 85.45**  & 0.73  & 85.05  & 0.75  & 94.58  & 85.44  & 65.29  & 87.74  & 95.05  & 82.30  & 88.76  & 85.55  & 87.49  & 74.48  & 88.92 \\
& \textbf{mean} & 85.48**  & 0.62  & 85.06  & 0.64  & 94.35  & 85.38  & 65.24  & 87.13  & 95.13  & 82.27  & 89.91  & 85.82  & 87.04  & 75.43  & 87.97 \\ \hline
\end{tabular}
}
\caption{GAT results of \textbf{AX} dataset, with each attention type and filters, averaged across 5 independent runs, numbers of layers (\{1, 2, 3\}) and hidden nodes (\{64, 128, 256\}). The name of the document classes are: Artificial Intelligence, Computational Engineering, Computer Vision, Data Structures, Information Theory, Neural and Evolutionary, Programming Languages, Systems and Control, Commutative Algebra, Group Theory, and Statistic Theory. * indicates a statistically significant improvement of the GAT accuracy over the SWA models ($p < 0.05$).  ** indicate higher improvement ($p < 0.01$).}
\label{ax-gat}
\end{table*}

\begin{table*}[]
\centering
\small
\begin{tabular}{ll|cccccc}
\hline
\multicolumn{1}{c}{\multirow{2}{*}{\textbf{Window}}} & \multicolumn{1}{c|}{\multirow{2}{*}{\textbf{Filter}}} & \multicolumn{1}{c}{\multirow{2}{*}{\textbf{Acc}}} & \multicolumn{1}{c}{\multirow{2}{*}{\textbf{Acc (std)}}} & \multicolumn{1}{c}{\multirow{2}{*}{\textbf{F1}}} & \multicolumn{1}{c}{\multirow{2}{*}{\textbf{F1 (std)}}} & \multicolumn{2}{c}{\textbf{F1 per class}} \\ \cline{7-8} 
\multicolumn{1}{c}{} & \multicolumn{1}{c|}{} & \multicolumn{1}{c}{} & \multicolumn{1}{c}{} & \multicolumn{1}{c}{} & \multicolumn{1}{c}{} & \textbf{non-hyperpartisan} & \textbf{hyperpartisan} \\ \hline
\multirow{3}{*}{\textbf{10}} & \textbf{unfilt} & 76.24** & 1.64 & 76.15 & 1.67 & 75.84 & 76.46 \\
 & \textbf{max} & 76.04* & 1.24 & 75.95 & 1.29 & 75.61 & 76.30 \\
 & \textbf{mean} & 75.59* & 1.66 & 75.47 & 1.77 & 75.63 & 75.31 \\ \hline
\multirow{3}{*}{\textbf{20}} & \textbf{unfilt} & 76.40** & 1.79 & 76.32 & 1.87 & 76.10 & 76.54 \\
 & \textbf{max} & 75.03 & 1.23 & 74.97 & 1.24 & 74.83 & 75.10 \\
 & \textbf{mean} & 75.34 & 1.59 & 75.27 & 1.61 & 75.16 & 75.39 \\ \hline
\multirow{3}{*}{\textbf{40}} & \textbf{unfilt} & 76.20* & 1.54 & 76.14 & 1.58 & 76.27 & 76.01 \\
 & \textbf{max} & 75.39 & 2.21 & 75.30 & 2.28 & 75.18 & 75.41 \\
 & \textbf{mean} & 75.50 & 1.57 & 75.44 & 1.62 & 75.83 & 75.05 \\ \hline
\multirow{3}{*}{\textbf{50}} & \textbf{unfilt} & 76.50 & 1.46 & 76.45 & 1.48 & 76.71 & 76.18 \\
 & \textbf{max} & 74.86 & 1.91 & 74.80 & 1.91 & 74.42 & 75.18 \\
 & \textbf{mean} & 75.31 & 1.36 & 75.23 & 1.37 & 74.85 & 75.61 \\ \hline
\end{tabular}
\caption{GAT results of \textbf{window size ablation study} on HND dataset, with each attention type and filters, averaged across 5 independent runs, numbers of layers (\{1, 2, 3\}) and hidden nodes (\{128, 256\}). * indicates a statistically significant improvement of the GAT accuracy over the SWA models ($p < 0.05$).  ** indicate higher improvement ($p < 0.01$).}
\label{wshnd-gat}
\end{table*}

\begin{table*}[]
\centering
\small
\begin{tabular}{ll|cccccc}
\hline
\multicolumn{1}{c}{\multirow{2}{*}{\textbf{Graph type}}} & \multicolumn{1}{c|}{\multirow{2}{*}{\textbf{Scheme}}} & \multicolumn{1}{c}{\multirow{2}{*}{\textbf{Acc}}} & \multicolumn{1}{c}{\multirow{2}{*}{\textbf{Acc (std)}}} & \multicolumn{1}{c}{\multirow{2}{*}{\textbf{F1}}} & \multicolumn{1}{c}{\multirow{2}{*}{\textbf{F1 (std)}}} & \multicolumn{2}{c}{\textbf{F1 per class}} \\ \cline{7-8} 
\multicolumn{1}{c}{} & \multicolumn{1}{c|}{} & \multicolumn{1}{c}{} & \multicolumn{1}{c}{} & \multicolumn{1}{c}{} & \multicolumn{1}{c}{} & \textbf{non-summary} & \textbf{summary} \\ \hline
\multirow{2}{*}{\textit{Baseline}} & \textbf{Order} & 61.04 &5.45 & 46.41 & 2.65 & 74.21 & 18.62 \\
& \textbf{Window} & 61.24 & 6.30 & 46.28 & 2.80 & 74.36 & 18.19 \\ \hline
\multirow{2}{*}{\textbf{Anneal}} & \textbf{max} & 57.21*  & 2.41  & 44.78  & 1.28  & 70.90  & 18.69 \\
 & \textbf{mean} & 53.96  & 3.12  & 42.69  & 1.71  & 67.91  & 17.48 \\ \hline
\multirow{2}{*}{\textbf{Softmax}} & \textbf{max} & 58.23*  & 2.80  & 45.36  & 1.46  & 71.76  & 18.98 \\
 & \textbf{mean} & 53.59  & 1.43  & 42.61  & 0.80  & 67.64  & 17.59\\ \hline
\multirow{2}{*}{\textbf{ReLU}} & \textbf{max} & 57.71  & 2.41  & 44.99  & 1.20  & 71.36  & 18.62 \\
 & \textbf{mean} & 55.22  & 1.87  & 43.44  & 0.98  & 69.09  & 17.76 \\ \hline
\multirow{2}{*}{\textbf{Sigmoid}} & \textbf{max} & 55.99  & 2.81  & 43.88  & 1.46  & 69.83  & 17.93 \\
 & \textbf{mean} & 54.08  & 3.18  & 42.57  & 1.63  & 68.06  & 17.06 \\ \hline
\end{tabular}
\caption{GAT results of \textbf{GR} dataset including baselines, with each attention type and filters, averaged across 5 independent runs, numbers of layers (\{1, 2, 3\}) and hidden nodes (\{64, 128, 256\}). * indicates a statistically significant improvement of the GAT accuracy over the SW models ($p < 0.05$).}
\label{gr-gat}
\end{table*}

\begin{table*}[t]
\centering
\scriptsize
\setlength{\tabcolsep}{2pt}
\renewcommand{\arraystretch}{1.1}
\emergencystretch=2em

\begin{tabularx}{\textwidth}{
l
>{\centering\arraybackslash}p{0.7cm}
>{\centering\arraybackslash}p{0.7cm}
>{\raggedright\arraybackslash}X
>{\centering\arraybackslash}p{0.7cm}
>{\centering\arraybackslash}p{0.7cm}
>{\raggedright\arraybackslash}X
}
\toprule
\multirow{2}{*}{\textbf{Config}} &
\multicolumn{3}{c}{\textbf{Acc}} &
\multicolumn{3}{c}{\textbf{F1}} \\
\cmidrule(lr){2-4}
\cmidrule(lr){5-7}
& \textbf{ANOVA} & \textbf{$p$} & \textbf{Groups}
& \textbf{ANOVA} & \textbf{$p$} & \textbf{Groups} \\
\midrule

Attention
& 0.397 & -- & --
& 0.200 & -- & -- \\

Filter
& \textless{}0.001 & 0.05 & --
& \textless{}0.001 & 0.05 & [max \textgreater mean] \\

& & 0.01 &
[0 \textgreater max], [0 \textgreater mean]
& & 0.01 &
[0 \textgreater max], [0 \textgreater mean] \\

Att + Filt
& \textless{}0.001 & 0.05 &
[Anneal\_0 \textgreater ReLU\_max],
[Anneal\_mean \textgreater ReLU\_0],
[Anneal\_mean \textgreater ReLU\_mean],
[Softmax\_0 \textgreater ReLU\_max],
[ReLU\_0 \textgreater ReLU\_max],
[ReLU\_max \textgreater ReLU\_mean]
& \textless{}0.001 & 0.05 &
[Anneal\_0 \textgreater Anneal\_max],
[Softmax\_0 \textgreater Anneal\_max] \\

& & 0.01 &
[Anneal\_0 \textgreater Anneal\_max],
[Anneal\_0 \textgreater Anneal\_mean],
[Anneal\_0 \textgreater ReLU\_mean],
[Anneal\_0 \textgreater Softmax\_max],
[Anneal\_0 \textgreater Softmax\_mean],
[Anneal\_0 \textgreater Sigmoid\_max],
[Anneal\_0 \textgreater Sigmoid\_mean],
[ReLU\_0 \textgreater Anneal\_max],
[ReLU\_0 \textgreater Softmax\_max],
[ReLU\_0 \textgreater Softmax\_mean],
[ReLU\_0 \textgreater ReLU\_mean],
[ReLU\_0 \textgreater Sigmoid\_max],
[ReLU\_0 \textgreater Sigmoid\_mean],
[Softmax\_0 \textgreater Anneal\_max],
[Softmax\_0 \textgreater ReLU\_mean],
[Softmax\_0 \textgreater Softmax\_max],
[Softmax\_0 \textgreater Softmax\_mean],
[Softmax\_0 \textgreater Sigmoid\_max],
[Softmax\_0 \textgreater Sigmoid\_mean],
[Sigmoid\_0 \textgreater Anneal\_max],
[Sigmoid\_0 \textgreater Anneal\_mean],
[Sigmoid\_0 \textgreater ReLU\_max],
[Sigmoid\_0 \textgreater ReLU\_mean],
[Sigmoid\_0 \textgreater Softmax\_max],
[Sigmoid\_0 \textgreater Softmax\_mean],
[Sigmoid\_0 \textgreater Sigmoid\_max],
[Sigmoid\_0 \textgreater Sigmoid\_mean]
& & 0.01 &
[Anneal\_0 \textgreater ReLU\_mean],
[Anneal\_0 \textgreater Softmax\_max],
[Anneal\_0 \textgreater Softmax\_mean],
[Anneal\_0 \textgreater Sigmoid\_max],
[Anneal\_0 \textgreater Sigmoid\_mean],
[Anneal\_max \textgreater ReLU\_mean],
[Anneal\_mean \textgreater ReLU\_mean],
[ReLU\_0 \textgreater ReLU\_mean],
[ReLU\_0 \textgreater Softmax\_max],
[ReLU\_0 \textgreater Softmax\_mean],
[ReLU\_0 \textgreater Sigmoid\_max],
[ReLU\_0 \textgreater Sigmoid\_mean],
[ReLU\_max \textgreater ReLU\_mean],
[Softmax\_0 \textgreater ReLU\_mean],
[Softmax\_0 \textgreater Softmax\_max],
[Softmax\_0 \textgreater Softmax\_mean],
[Softmax\_0 \textgreater Sigmoid\_max],
[Softmax\_0 \textgreater Sigmoid\_mean],
[Sigmoid\_0 \textgreater Anneal\_max],
[Sigmoid\_0 \textgreater Anneal\_mean],
[Sigmoid\_0 \textgreater ReLU\_max],
[Sigmoid\_0 \textgreater ReLU\_mean],
[Sigmoid\_0 \textgreater Softmax\_max],
[Sigmoid\_0 \textgreater Softmax\_mean],
[Sigmoid\_0 \textgreater Sigmoid\_max],
[Sigmoid\_0 \textgreater Sigmoid\_mean] \\

Att + Layer
& 0.358 & -- & --
& 0.182 & -- & -- \\

Att + Hidden
& 0.014 & 0.05 &
[Anneal\_64 \textgreater ReLU\_256],
[Anneal\_256 \textgreater ReLU\_64]
& 0.014 & 0.05 & -- \\

& & 0.01 & --
& & 0.01 &
[Anneal\_64 \textgreater ReLU\_256] \\

Filt + Layer
& \textless{}0.001 & 0.05 &
[0\_2 \textgreater 0\_1],
[0\_1 \textgreater max\_3]
& \textless{}0.001 & 0.05 &
[0\_2 \textgreater 0\_1],
[0\_3 \textgreater max\_3],
[0\_1 \textgreater max\_2] \\

& & 0.01 &
[0\_1 \textgreater max\_1], [0\_1 \textgreater max\_2],
[0\_1 \textgreater mean\_1], [0\_1 \textgreater mean\_2],
[0\_1 \textgreater mean\_3], [0\_2 \textgreater max\_1],
[0\_2 \textgreater max\_2], [0\_2 \textgreater max\_3],
[0\_2 \textgreater mean\_1], [0\_2 \textgreater mean\_2],
[0\_2 \textgreater mean\_3], [0\_3 \textgreater max\_1],
[0\_3 \textgreater max\_2], [0\_3 \textgreater max\_3],
[0\_3 \textgreater mean\_1], [0\_3 \textgreater mean\_2],
[0\_3 \textgreater mean\_3]
& & 0.01 &
[0\_1 \textgreater max\_1], [0\_1 \textgreater mean\_3],
[0\_3 \textgreater max\_2], [0\_3 \textgreater max\_1],
[0\_1 \textgreater mean\_2], [0\_1 \textgreater mean\_1],
[0\_3 \textgreater mean\_3], [0\_3 \textgreater mean\_2],
[0\_2 \textgreater max\_3], [0\_3 \textgreater mean\_1],
[0\_2 \textgreater max\_2], [0\_2 \textgreater max\_1],
[0\_2 \textgreater mean\_3], [0\_2 \textgreater mean\_2],
[0\_2 \textgreater mean\_1] \\

Filt + Hidden
& \textless{}0.001 & 0.05 &
[0\_256 \textgreater max\_128],
[0\_256 \textgreater max\_64],
[0\_64 \textgreater 0\_256],
[max\_64 \textgreater mean\_256]
& \textless{}0.001 & 0.05 &
[0\_256 \textgreater max\_256],
[max\_256 \textgreater mean\_256],
[max\_64 \textgreater mean\_256] \\

& & 0.01 &
[0\_128 \textgreater max\_128],
[0\_128 \textgreater max\_256],
[0\_128 \textgreater max\_64],
[0\_128 \textgreater mean\_128],
[0\_128 \textgreater mean\_256],
[0\_128 \textgreater mean\_64],
[0\_256 \textgreater max\_256],
[0\_256 \textgreater mean\_128],
[0\_256 \textgreater mean\_256],
[0\_64 \textgreater max\_128],
[0\_64 \textgreater max\_256],
[0\_64 \textgreater max\_64],
[0\_64 \textgreater mean\_128],
[0\_64 \textgreater mean\_256],
[0\_64 \textgreater mean\_64],
[max\_128 \textgreater mean\_256],
[mean\_64 \textgreater mean\_256]
& & 0.01 &
[0\_128 \textgreater max\_128],
[0\_128 \textgreater max\_256],
[0\_128 \textgreater max\_64],
[0\_128 \textgreater mean\_128],
[0\_128 \textgreater mean\_256],
[0\_128 \textgreater mean\_64],
[0\_256 \textgreater mean\_128],
[0\_256 \textgreater mean\_256],
[0\_64 \textgreater max\_128],
[0\_64 \textgreater max\_256],
[0\_64 \textgreater max\_64],
[0\_64 \textgreater mean\_128],
[0\_64 \textgreater mean\_256],
[0\_64 \textgreater mean\_64],
[max\_128 \textgreater mean\_256],
[mean\_64 \textgreater mean\_256] \\

\bottomrule
\end{tabularx}

\caption{ANOVA results and statistically significant pairwise group differences for each configuration factor under Accuracy (Acc) and F1 score in \textbf{BBC} dataset. 0 refers to unfiltered graphs.}
\label{tab:bbc-stat-ana}
\end{table*}

\begin{table*}[t]
\centering
\scriptsize
\setlength{\tabcolsep}{2pt}
\renewcommand{\arraystretch}{1.1}
\emergencystretch=2em

\begin{tabularx}{\textwidth}{
l
>{\centering\arraybackslash}p{0.7cm}
>{\centering\arraybackslash}p{0.7cm}
>{\raggedright\arraybackslash}X
>{\centering\arraybackslash}p{0.7cm}
>{\centering\arraybackslash}p{0.7cm}
>{\raggedright\arraybackslash}X
}
\toprule
\multirow{2}{*}{\textbf{Config}} &
\multicolumn{3}{c}{\textbf{Acc}} &
\multicolumn{3}{c}{\textbf{F1}} \\
\cmidrule(lr){2-4}
\cmidrule(lr){5-7}
& \textbf{ANOVA} & \textbf{$p$} & \textbf{Groups}
& \textbf{ANOVA} & \textbf{$p$} & \textbf{Groups} \\
\midrule

Attention
& \textless{}0.001 & 0.01 &
[ReLU \textgreater Softmax], [Sigmoid \textgreater Softmax]
& \textless{}0.001 & 0.01 &
[ReLU \textgreater Softmax], [Sigmoid \textgreater Softmax] \\

Filter
& 0.002 & 0.05 & --
& 0.004 & 0.05 &
[mean \textgreater max] \\

& & 0.01 &
[mean \textgreater max]
& & 0.01 & -- \\

Att + Filt
& \textless{}0.001 & 0.05 &
[Anneal\_mean \textgreater Anneal\_0],
[Sigmoid\_max \textgreater Softmax\_mean]
& \textless{}0.001 & 0.05 &
[Anneal\_mean \textgreater Anneal\_0],
[Sigmoid\_max \textgreater Softmax\_mean] \\

& & 0.01 &
[ReLU\_0 \textgreater Softmax\_max],
[ReLU\_max \textgreater Softmax\_max],
[ReLU\_mean \textgreater Softmax\_max],
[Softmax\_mean \textgreater Softmax\_max],
[Sigmoid\_0 \textgreater Softmax\_max],
[Sigmoid\_max \textgreater Softmax\_max],
[Sigmoid\_mean \textgreater Softmax\_max]
& & 0.01 &
[ReLU\_0 \textgreater Softmax\_max],
[ReLU\_max \textgreater Softmax\_max],
[ReLU\_mean \textgreater Softmax\_max],
[Softmax\_mean \textgreater Softmax\_max],
[Sigmoid\_0 \textgreater Softmax\_max],
[Sigmoid\_max \textgreater Softmax\_max],
[Sigmoid\_max \textgreater Softmax\_mean],
[Sigmoid\_mean \textgreater Softmax\_max] \\

Att + Layer
& \textless{}0.001 & 0.05 &
[Anneal\_3 \textgreater Anneal\_1],
[Softmax\_3 \textgreater Softmax\_1],
[ReLU\_2 \textgreater Anneal\_1],
[Sigmoid\_2 \textgreater Softmax\_3],
[Sigmoid\_2 \textgreater Sigmoid\_1],
[Sigmoid\_3 \textgreater Anneal\_1]
& \textless{}0.001 & 0.05 &
[Anneal\_3 \textgreater Anneal\_1],
[Softmax\_3 \textgreater Softmax\_1],
[ReLU\_2 \textgreater Anneal\_1],
[ReLU\_3 \textgreater Softmax\_3],
[Sigmoid\_1 \textgreater Softmax\_1],
[Sigmoid\_2 \textgreater Softmax\_3],
[Sigmoid\_3 \textgreater Anneal\_1],
[Sigmoid\_3 \textgreater ReLU\_1] \\

& & 0.01 &
[Anneal\_2 \textgreater Anneal\_1],
[ReLU\_2 \textgreater Softmax\_1],
[ReLU\_2 \textgreater ReLU\_1],
[ReLU\_3 \textgreater Anneal\_1],
[ReLU\_3 \textgreater Softmax\_1],
[ReLU\_3 \textgreater Softmax\_3],
[ReLU\_3 \textgreater ReLU\_1],
[Softmax\_2 \textgreater Softmax\_1],
[Sigmoid\_1 \textgreater Softmax\_1],
[Sigmoid\_2 \textgreater Anneal\_1],
[Sigmoid\_2 \textgreater ReLU\_1],
[Sigmoid\_2 \textgreater Softmax\_1],
[Sigmoid\_3 \textgreater ReLU\_1],
[Sigmoid\_3 \textgreater Softmax\_1]
& & 0.01 &
[Anneal\_2 \textgreater Anneal\_1],
[ReLU\_2 \textgreater ReLU\_1],
[ReLU\_2 \textgreater Softmax\_1],
[ReLU\_3 \textgreater Anneal\_1],
[ReLU\_3 \textgreater ReLU\_1],
[ReLU\_3 \textgreater Softmax\_1],
[Softmax\_2 \textgreater Softmax\_1],
[Sigmoid\_2 \textgreater Anneal\_1],
[Sigmoid\_2 \textgreater ReLU\_1],
[Sigmoid\_2 \textgreater Softmax\_1],
[Sigmoid\_3 \textgreater Softmax\_1] \\

Att + Hidden
& \textless{}0.001 & 0.05 &
[Sigmoid\_64 \textgreater Softmax\_256],
[ReLU\_64 \textgreater Softmax\_128]
& \textless{}0.001 & 0.05 &
[Sigmoid\_64 \textgreater Softmax\_256],
[ReLU\_64 \textgreater Softmax\_128] \\

& & 0.01 &
[Sigmoid\_128 \textgreater Softmax\_128],
[Sigmoid\_128 \textgreater Softmax\_256],
[Sigmoid\_128 \textgreater Softmax\_64],
[Sigmoid\_64 \textgreater Softmax\_128]
& & 0.01 &
[Sigmoid\_128 \textgreater Softmax\_128],
[Sigmoid\_128 \textgreater Softmax\_256] \\

Filt + Layer
& \textless{}0.001 & 0.05 & --
& \textless{}0.001 & 0.05 &
[max\_2 \textgreater 0\_1] \\

& & 0.01 &
[0\_3 \textgreater 0\_1],
[max\_2 \textgreater 0\_1],
[mean\_2 \textgreater 0\_1],
[max\_2 \textgreater max\_1],
[max\_3 \textgreater max\_1],
[mean\_1 \textgreater max\_1],
[mean\_2 \textgreater max\_1],
[mean\_3 \textgreater max\_1],
[mean\_2 \textgreater mean\_1]
& & 0.01 &
[mean\_2 \textgreater max\_1],
[max\_2 \textgreater max\_1],
[mean\_3 \textgreater max\_1],
[max\_3 \textgreater max\_1],
[0\_3 \textgreater 0\_1],
[mean\_2 \textgreater 0\_1],
[mean\_1 \textgreater max\_1],
[mean\_2 \textgreater mean\_1] \\

Filt + Hidden
& 0.030 & 0.05 &
[mean\_64 \textgreater max\_128]
& 0.030 & 0.05 & -- \\

\bottomrule
\end{tabularx}

\caption{ANOVA results and statistically significant pairwise group differences for each configuration factor under Accuracy (Acc) and F1 score in the \textbf{HND} dataset. 0 refers to unfiltered graphs.}
\label{tab:hnd-stat-ana}
\end{table*}

\begin{table*}[t]
\centering
\scriptsize
\setlength{\tabcolsep}{2pt}
\renewcommand{\arraystretch}{1.1}
\emergencystretch=2em

\begin{tabularx}{\textwidth}{
l
>{\centering\arraybackslash}p{0.7cm}
>{\centering\arraybackslash}p{0.7cm}
>{\raggedright\arraybackslash}X
>{\centering\arraybackslash}p{0.7cm}
>{\centering\arraybackslash}p{0.7cm}
>{\raggedright\arraybackslash}X
}
\toprule
\multirow{2}{*}{\textbf{Config}} &
\multicolumn{3}{c}{\textbf{Acc}} &
\multicolumn{3}{c}{\textbf{F1}} \\
\cmidrule(lr){2-4}
\cmidrule(lr){5-7}
& \textbf{ANOVA} & \textbf{$p$} & \textbf{Groups}
& \textbf{ANOVA} & \textbf{$p$} & \textbf{Groups} \\
\midrule

Attention
& 0.459 & -- & --
& 0.355 & -- & -- \\

Filter
& 0.340 & -- & --
& 0.359 & -- & -- \\

Att + Filt
& 0.378 & -- & --
& 0.450 & -- & -- \\

Att + Layer
& \textless{}0.001 & 0.01 &
[ReLU\_1 \textgreater ReLU\_2], [ReLU\_3 \textgreater Softmax\_1], [ReLU\_2 \textgreater Softmax\_1],
[ReLU\_3 \textgreater ReLU\_1], [Softmax\_3 \textgreater ReLU\_1],
[Softmax\_2 \textgreater ReLU\_1], [Softmax\_2 \textgreater Softmax\_1],
[Softmax\_3 \textgreater Softmax\_1]
& \textless{}0.001 & 0.01 &
[ReLU\_2 \textgreater ReLU\_1], [ReLU\_2 \textgreater Softmax\_1],
[ReLU\_3 \textgreater ReLU\_1], [ReLU\_3 \textgreater Softmax\_1],
[Softmax\_2 \textgreater ReLU\_1], [Softmax\_2 \textgreater Softmax\_1],
[Softmax\_3 \textgreater ReLU\_1], [Softmax\_3 \textgreater Softmax\_1] \\

Att + Hidden
& 0.438 & -- & --
& 0.405 & -- & -- \\

Filt + Layer
& \textless{}0.001 & 0.01 &
[max\_2 \textgreater max\_1], [max\_2 \textgreater mean\_1], [max\_3 \textgreater max\_1],
[max\_3 \textgreater mean\_1], [mean\_2 \textgreater max\_1],
[mean\_2 \textgreater mean\_1], [mean\_3 \textgreater max\_1],
[mean\_3 \textgreater mean\_1]
& \textless{}0.001 & 0.01 &
[max\_2 \textgreater max\_1], [max\_2 \textgreater mean\_1], [max\_3 \textgreater max\_1],
[max\_3 \textgreater mean\_1], [mean\_2 \textgreater max\_1],
[mean\_2 \textgreater mean\_1], [mean\_3 \textgreater max\_1],
[mean\_3 \textgreater mean\_1] \\

Filt + Hidden
& 0.674 & -- & --
& 0.650 & -- & -- \\

\bottomrule
\end{tabularx}

\caption{ANOVA results and statistically significant pairwise group differences for each configuration factor under Accuracy (Acc) and F1 score in \textbf{AX} dataset. 0 refers to unfiltered graphs.}
\label{tab:ax-stat-ana}
\end{table*}

\begin{table*}[t]
\centering
\scriptsize
\setlength{\tabcolsep}{2pt}
\renewcommand{\arraystretch}{1.1}
\emergencystretch=2em

\begin{tabularx}{\textwidth}{
l
>{\centering\arraybackslash}p{0.7cm}
>{\centering\arraybackslash}p{0.7cm}
>{\raggedright\arraybackslash}X
>{\centering\arraybackslash}p{0.7cm}
>{\centering\arraybackslash}p{0.7cm}
>{\raggedright\arraybackslash}X
}
\toprule
\multirow{2}{*}{\textbf{Config}} &
\multicolumn{3}{c}{\textbf{Acc}} &
\multicolumn{3}{c}{\textbf{F1}} \\
\cmidrule(lr){2-4}
\cmidrule(lr){5-7}
& \textbf{ANOVA} & \textbf{$p$} & \textbf{Groups}
& \textbf{ANOVA} & \textbf{$p$} & \textbf{Groups} \\
\midrule

Attention
& 0.189 & -- & --
& 0.045 & 0.05 &
[ReLU \textgreater Sigmoid] \\

Filter
& \textless{}0.001 & 0.01 &
[max \textgreater mean]
& \textless{}0.001 & 0.01 &
[max \textgreater mean] \\

Att + Filt
& \textless{}0.001 & 0.05 &
[Anneal\_max \textgreater Sigmoid\_mean],
[Softmax\_max \textgreater ReLU\_mean]
& \textless{}0.001 & 0.05 &
[ReLU\_max \textgreater ReLU\_mean],
[Softmax\_max \textgreater Sigmoid\_max] \\

& & 0.01 &
[Anneal\_max \textgreater Anneal\_mean],
[Anneal\_max \textgreater Softmax\_mean],
[ReLU\_max \textgreater Anneal\_mean],
[ReLU\_max \textgreater Softmax\_mean],
[ReLU\_max \textgreater Sigmoid\_mean],
[Softmax\_max \textgreater Sigmoid\_mean],
[Softmax\_max \textgreater Softmax\_mean],
[Softmax\_max \textgreater Anneal\_mean]
& & 0.01 &
[Anneal\_max \textgreater Anneal\_mean],
[Anneal\_max \textgreater Softmax\_mean],
[Anneal\_max \textgreater Sigmoid\_mean],
[ReLU\_max \textgreater Anneal\_mean],
[ReLU\_max \textgreater Softmax\_mean],
[ReLU\_max \textgreater Sigmoid\_mean],
[Softmax\_max \textgreater Anneal\_mean],
[Softmax\_max \textgreater ReLU\_mean],
[Softmax\_max \textgreater Softmax\_mean],
[Softmax\_max \textgreater Sigmoid\_mean] \\

Att + Layer
& 0.821 & -- & --
& 0.523 & -- & -- \\

Att + Hidden
& 0.002 & 0.05 &
[ReLU\_256 \textgreater Sigmoid\_128]
& 0.001 & 0.05 &
[Softmax\_256 \textgreater Sigmoid\_128],
[ReLU\_256 \textgreater Sigmoid\_128] \\

Filt + Layer
& \textless{}0.001 & 0.05 &
[max\_1 \textgreater mean\_3],
[max\_3 \textgreater mean\_1]
& \textless{}0.001 & 0.05 &
[max\_3 \textgreater mean\_3] \\

& & 0.01 &
[max\_2 \textgreater mean\_3],
[max\_1 \textgreater mean\_1],
[max\_3 \textgreater mean\_2],
[max\_2 \textgreater mean\_1],
[max\_1 \textgreater mean\_2],
[max\_2 \textgreater mean\_2]
& & 0.01 &
[max\_1 \textgreater mean\_1],
[max\_1 \textgreater mean\_2],
[max\_1 \textgreater mean\_3],
[max\_2 \textgreater mean\_1],
[max\_2 \textgreater mean\_2],
[max\_2 \textgreater mean\_3],
[max\_3 \textgreater mean\_1],
[max\_3 \textgreater mean\_2] \\

Filt + Hidden
& \textless{}0.001 & 0.05 &
[max\_128 \textgreater mean\_256],
[max\_128 \textgreater mean\_64]
& \textless{}0.001 & 0.05 &
[max\_128 \textgreater mean\_256],
[max\_128 \textgreater mean\_64] \\

& & 0.01 &
[max\_128 \textgreater mean\_128],
[max\_256 \textgreater max\_64],
[max\_256 \textgreater mean\_256],
[max\_256 \textgreater mean\_64],
[max\_256 \textgreater mean\_128],
[max\_256 \textgreater max\_128]
& & 0.01 &
[max\_128 \textgreater mean\_256],
[max\_128 \textgreater mean\_64],
[max\_128 \textgreater mean\_128],
[max\_256 \textgreater max\_64],
[max\_256 \textgreater mean\_256],
[max\_256 \textgreater mean\_64],
[max\_256 \textgreater mean\_128],
[max\_256 \textgreater max\_128] \\

\bottomrule
\end{tabularx}

\caption{ANOVA results and statistically significant pairwise group differences for each configuration factor under Accuracy (Acc) and F1 score in the \textbf{GR} dataset. 0 refers to unfiltered graphs.}
\label{tab:gr-stat-ana}
\end{table*}

\begin{figure*}[t]
  \includegraphics[width=\textwidth]{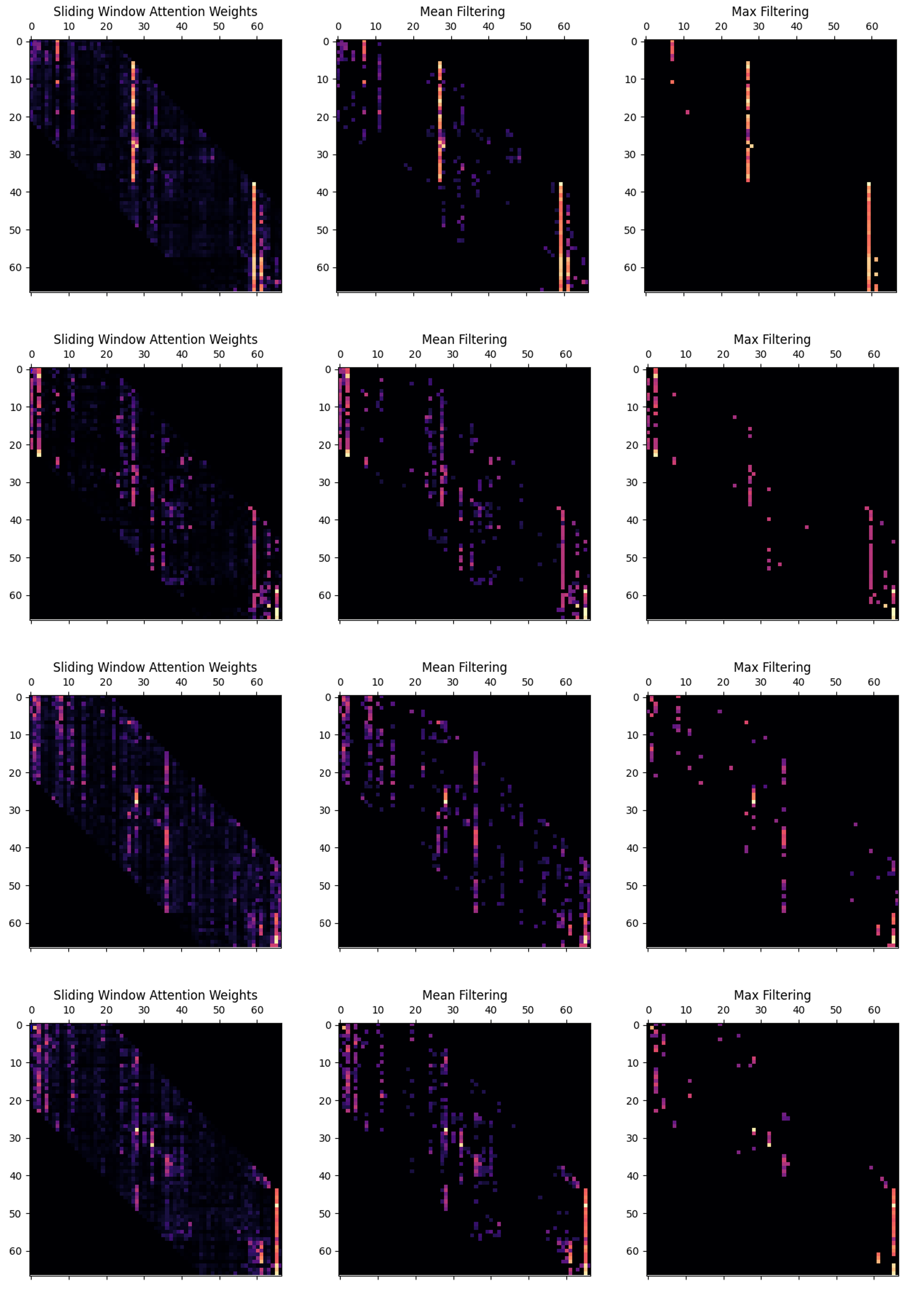}
  \caption{A sample of attention matrices from a randomly selected document in the GovReport dataset. From Top to Bottom: conventional Softmax, Annealing Softmax, ReLU, and Sigmoid. Brighter colors illustrate higher values.}
  \label{fig:attn}
\end{figure*}

\section{Investigation into Extractive Summarization}
\label{sec:exp}

The predicted summaries for this investigation are produced using the prediction on the test split. We used the best model, namely, the max-bound ReLU-attention two-layer GAT with 256 hidden nodes, which achieves the highest F1 score of 0.51 on the validation split.

\subsection{Summary Sentence Distribution}
\label{sec:gr-sentdist}

\begin{figure*}[t]
  \includegraphics[width=\textwidth]{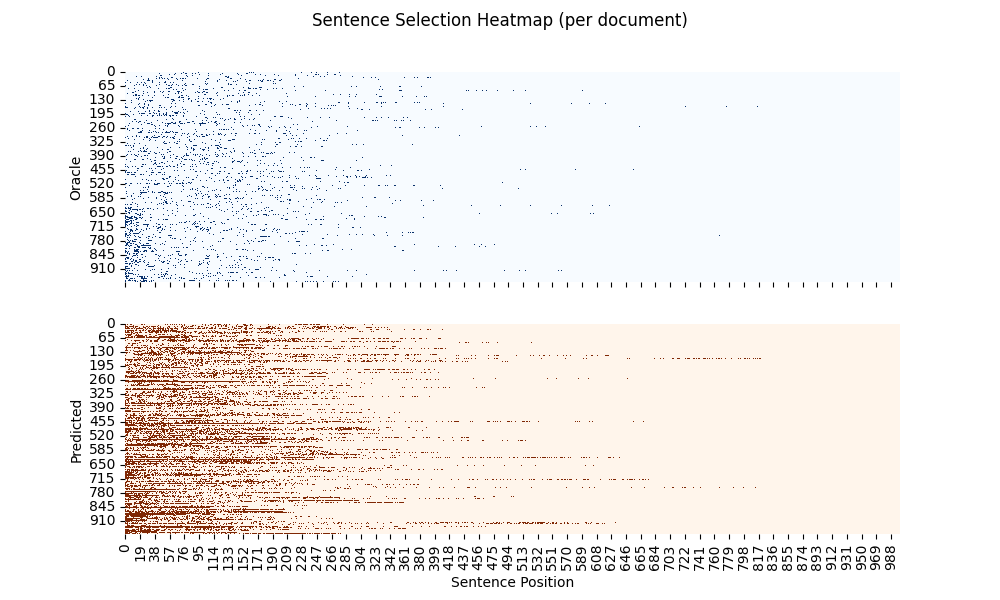}
  \caption{The Figure shows the occurrences of summary sentences from random documents in the test split. Note that the sentence positions are absolute positions with padded length to the 1000th position.}
  \label{fig:heatsum}
\end{figure*}

Assuming GAT models might benefit from the order of the sentences, we checked if there is a tendency that a summary sentence is likely to appear in any relative position. 
The findings indicate only a weak positional trend that earlier sentences are somewhat more likely to appear in summaries, whereas later ones are less likely.
Beyond this, however, we did not observe any systematic positional bias in either the oracle or the predicted summaries.
Summary sentences are shown to spread throughout the entire document (see \autoref{fig:heatsum}).

Although we could not prove that there is a relative order of summary sentences, we also could not rule out the possibility that heuristic graphs preserve certain information related to sentence order that is overlooked by attention-based graphs.

\begin{figure*}[t]
  \includegraphics[width=\textwidth]{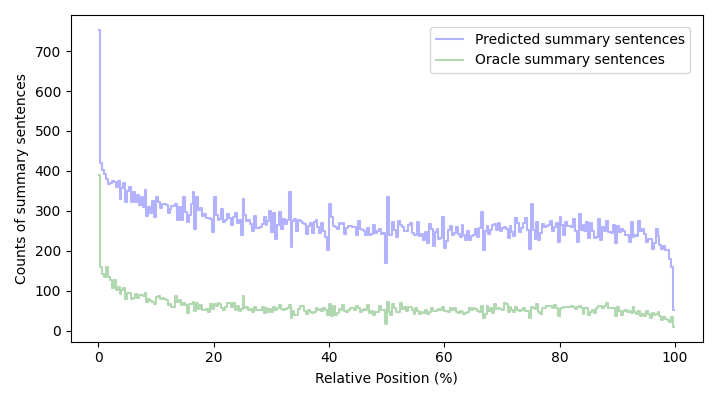}
  \caption{The figure shows the distribution of the summary sentences in relative positions (\%) of the documents in the test split. Predicted summary sentences are shown in purple, and oracle summary sentences in green.}
  \label{fig:sentpostest}
\end{figure*}

\subsection{Ratio of Summary Sentences}
\label{sec:gr-sentratio}

\begin{figure*}[!htbp]
  \includegraphics[width=0.48\linewidth]{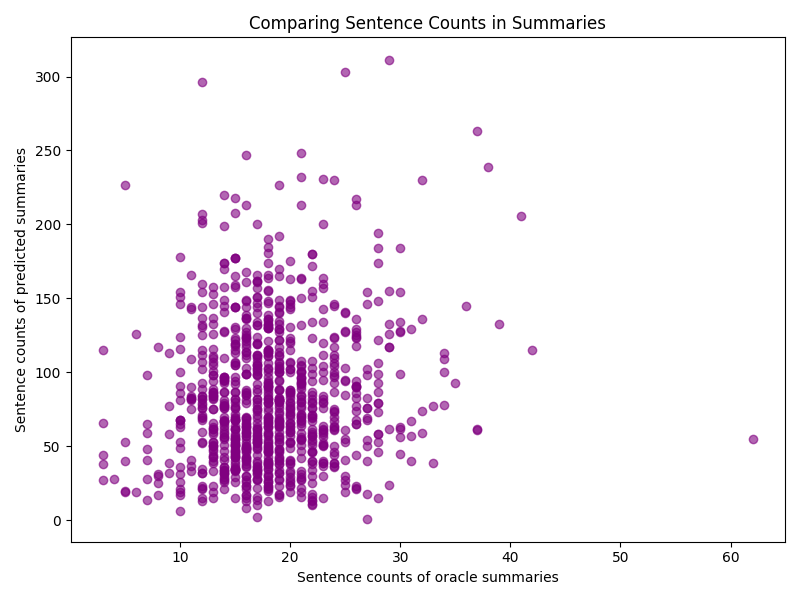} \hfill
  \includegraphics[width=0.48\linewidth]{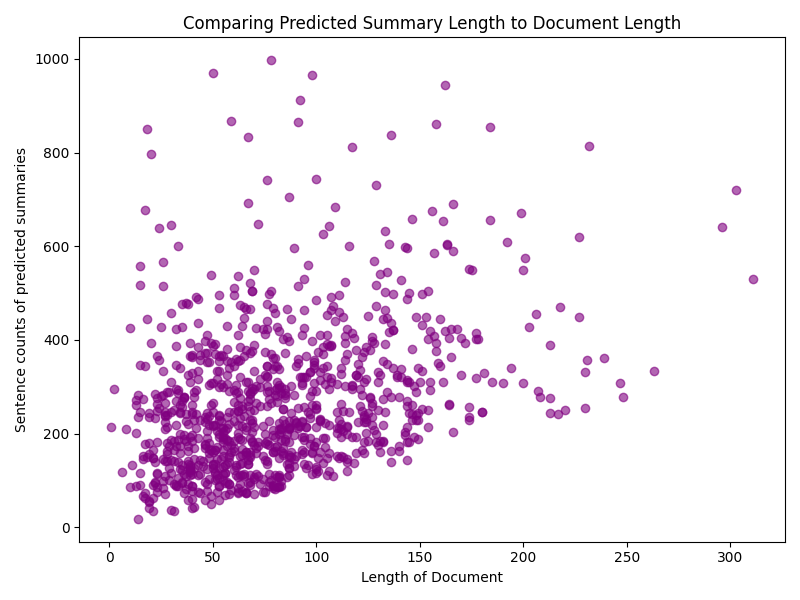}
  \caption {Left: The numbers of Oracle summary sentences (y-axis), compared to the numbers of predicted summary sentences (x-axis). Right: The length of documents (y-axis), compared to the number of predicted summary sentences (x-axis).}
  \label{fig:psum}
\end{figure*}

The result of this investigation exhibits the tendency of the GAT model to over-predict sentences to be a part of summaries.
While an oracle summary includes an average of 6\% of sentences in a document, the model predicts that roughly 20\% of sentences should be included (see \autoref{sec:gr-sentdist}).
It also shows exponential trends to include more sentences as a part of the summary when a document is longer (see \autoref{fig:psum}).

A strong bias towards including sentences in the summary may result from a weighted loss function, model parameter settings, or even the ordering of the training samples.

\subsection{Similarity Scores}
\label{sec:gr-simscore}

We evaluated the predicted summaries using two qualitative text similarity scores, namely ROUGE scores \cite{lin2004rouge} and BERTscore \cite{zhang2019bertscore}.
Primarily, ROUGE scores are keyword-based, while BERTScore focuses on semantic similarity but minimal syntactic similarity.

\begin{figure*}[t]
  \includegraphics[width=0.5\linewidth]{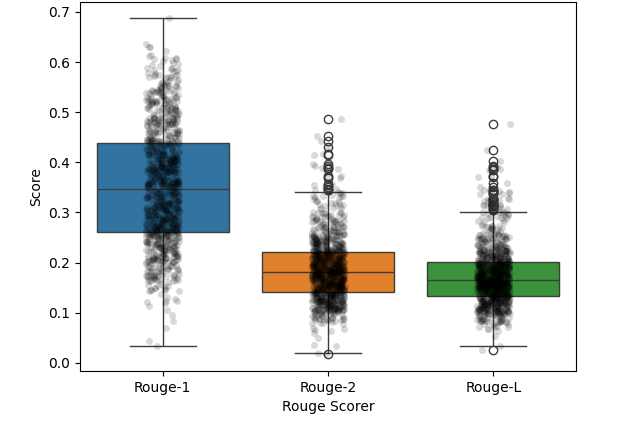} \hfill
  \includegraphics[width=0.5\linewidth]{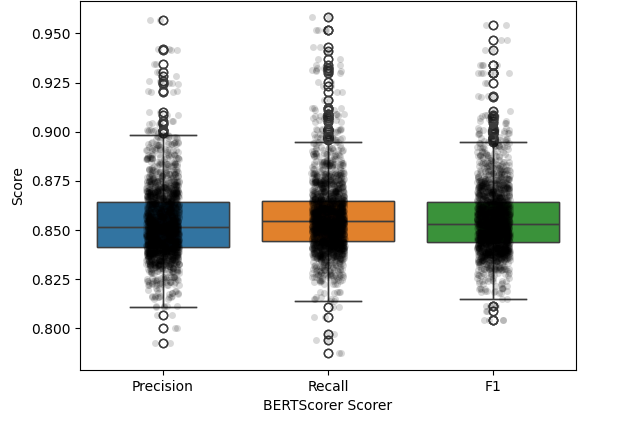}
  \caption {Text similarity scores on summaries generated from the single-layer GAT model trained on ReLU-based graphs. Left: ROUGE-1, ROUGE-2, and ROUGE-L F1 scores on the test split. Right: BERTScore in terms of precision, recall, and F1.}
  \label{fig:simscore}
\end{figure*}

Three ROUGE scores used in this experiment are: ROUGE-1, ROUGE-2, and ROUGE-L; in other words, the overlaps of unigrams, bigrams, and the longest common subsequence, respectively (see \autoref{fig:simscore}).
Considering both the previous F1 score and the current ROUGE-1 score of 0.35, indicates that the model picks an acceptable number of the oracle sentences, but it might not capture keywords in the exact surface form.

On the other hand, achieving the BERTScore of 0.83 shows that the predicted summaries retain high semantic similarity to the gold summaries.
Moreover, a higher ratio of summary sentences (see \autoref{sec:gr-sentratio}) suggests that GAT models prefer to greedily cover a broader set of sentences that are semantically aligned with the actual summaries, rather than a handful of densely packed sentences with keywords as in oracle summaries.
In other words, the models appear to favor semantic coverage and topic coherence over purely maximizing lexical overlap.

\subsection{t-SNE Representation Analysis}

\begin{figure*}[t]
    \centering
    \includegraphics[width=0.8\textwidth]{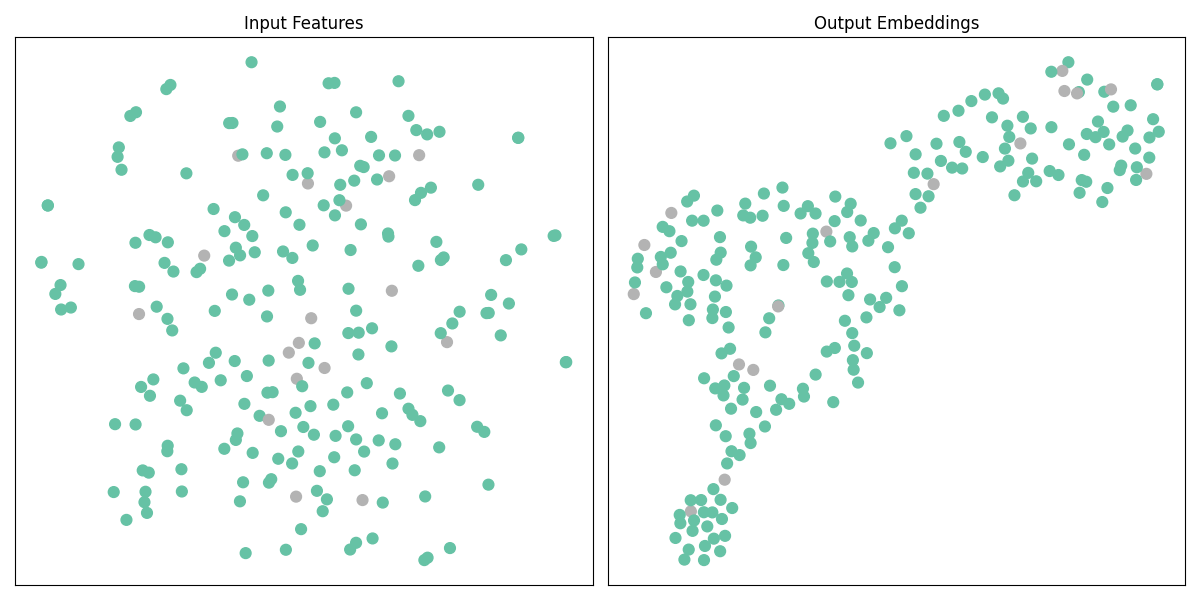}
    \vspace{0.5em}
    \includegraphics[width=0.8\textwidth]{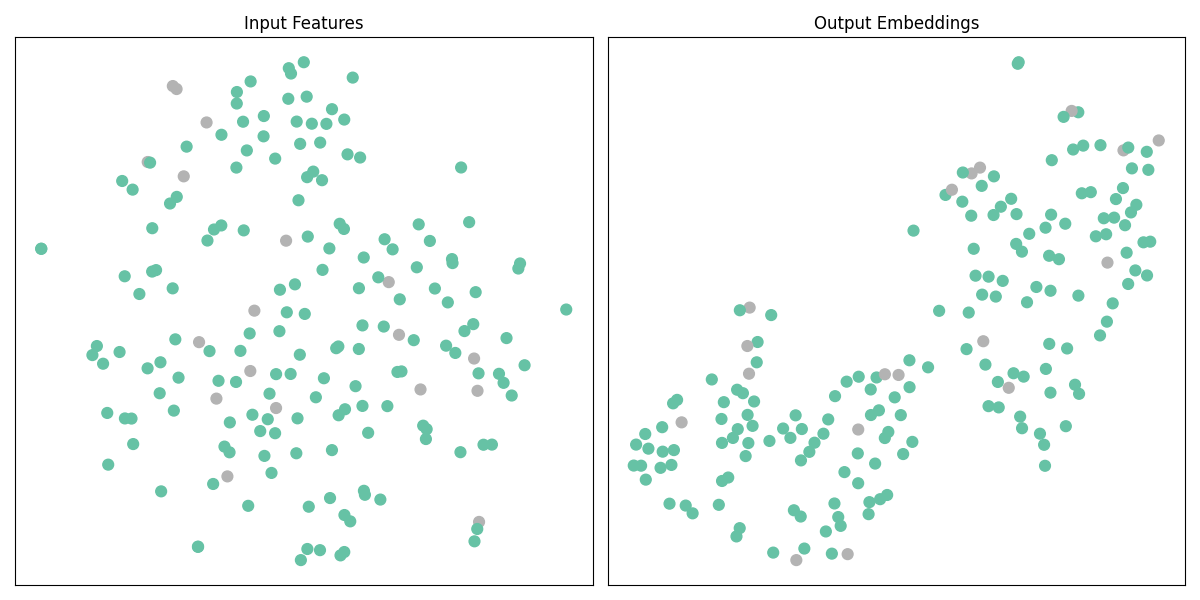}
    \caption{t-SNE visualizations of random GR documents. Left-side images show the input feature representations, and the right show the learned output embeddings. Gray dots indicate sentences selected by the oracle summary.}
    \label{fig:tsne-samples}
\end{figure*}

\autoref{fig:tsne-samples} shows examples of t-SNE projection of the sentence embeddings. 
Sentences selected for the reference summaries tend to appear near the centers of broader sentence clusters rather than forming clearly separated regions. 
Consistent with intuition, summary-worthy sentences often represent the central themes of a document section.

The visualization also helps explain the difficulty of the sentence classification task,
Numerous non-summary sentences lie close to summary sentences and convey overlapping information.
This observation supports the interpretation in \autoref{sec:res-summ} that the model captures meaningful semantic centrality, yet this might not fully correspond to the evaluation against oracle summaries.

\end{document}